\documentclass[runningheads]{llncs}

\usepackage{eccv}

\usepackage{eccvabbrv}

\usepackage{graphicx}
\usepackage{booktabs}
\usepackage{makecell}

\usepackage{xparse}
\usepackage{multirow}
\usepackage[table]{xcolor}
\usepackage{xcolor}
\usepackage{soul}
\definecolor{blue}{HTML}{ecf9fd}
\definecolor{gray}{HTML}{f4f4f4}
\definecolor{yellow}{HTML}{fefbe1}
\definecolor{green}{HTML}{e1fee4}
\usepackage{graphicx}
\usepackage{wrapfig}
\usepackage{arydshln}
\usepackage[table]{xcolor}
\usepackage{pifont}
\newcommand{\cmark}{\ding{51}}
\newcommand{\xmark}{\ding{55}}

\usepackage{enumitem}

\newcommand{\method}{TRAG}

\usepackage[accsupp]{axessibility}  

\usepackage[pagebackref,breaklinks,colorlinks,citecolor=eccvblue]{hyperref}

\usepackage{orcidlink}

\begin{document}

\title{Token-level Response-visual Attention Guidance for Multimodal LLMs Knowledge Distillation} 

\titlerunning{Token-level Response-visual Attention Guidance for MLLMs KD}

\author{Jaehyun Jang\inst{1}\orcidlink{0009-0007-7622-857X} \and
Eunseop Yoon\inst{1}\orcidlink{0000-0002-5580-5354} \and
Hee Suk Yoon\inst{1}\orcidlink{0000-0003-2115-8459} \and
SooHwan Eom\inst{1}\orcidlink{0009-0007-7076-9471} \and \\
Mark A. Hasegawa-Johnson\inst{2}\orcidlink{0000-0002-5631-2893} \and
Chang D. Yoo\inst{1}\thanks{Corresponding Author}\orcidlink{0000-0002-0756-7179}
}

\authorrunning{J.~Jang et al.}

\institute{Korea Advanced Institute of Science and Technology, Daejeon,
Republic of Korea\\
\email{\{jhjangjh,esyoon97,hskyoon,sean1105,cd\_yoo\}@kaist.ac.kr}
 \and
University of Illinois Urbana-Champaign, Champaign, USA \\
\email{jhasegaw@illinois.edu}}

\maketitle

\begin{abstract}
    While knowledge distillation (KD) is widely adopted for training lightweight models by leveraging supervision from larger teacher models, relying solely on output token distributions has proven insufficient for compressing Multimodal Large Language Models (MLLMs). Since output tokens are a byproduct of the model attending to visual inputs, prior works have explored explicitly distilling attention to provide a direct supervisory signal. While promising, the precise utility of which attention signals to distill remains under-explored. In this work, we challenge the conventional reliance on prompt-to-vision attention by revealing that downstream performance correlates strongly with response-to-vision attention similarity to the teacher, but negligibly with that of prompt-conditioned attention. Furthermore, we observe that attention distributions exhibit significant variance across individual tokens, indicating that a uniform distillation objective is suboptimal. To this end, we introduce \textbf{\underline{T}oken-level \underline{R}esponse-visual \underline{A}ttention \underline{G}uidance (TRAG)}, a distillation objective that 1) shifts the focus to response-to-vision signals and 2) employs token-specific objectives by adaptively weighting the Kullback-Leibler divergence based on attention entropy, effectively guiding the student to mirror the teacher's precise visual focus. Extensive experimental results on multiple benchmarks demonstrate that TRAG significantly outperforms prior distillation baselines. Our code is available at \url{https://github.com/jhjangjh/TRAG}.
\end{abstract}

\section{Introduction} \label{sec:intro}

Multimodal Large Language Models (MLLMs)~\cite{llava, vila, deepseekvl, qwen3vl} have achieved remarkable performance, yet the continuous scaling of their architectures and datasets entails prohibitive computational costs. This deployment bottleneck has intensified research into MLLM compression~\cite{qvlm, vispruner}, where knowledge distillation (KD)~\cite{hintonkd} has emerged as a predominant paradigm, building upon its established effectiveness in standard LLM frameworks~\cite{minillm, gkd, distillm}. However, standard distillation objectives designed for text-only LLMs rely solely on output distributions and thus do not explicitly transfer how visual evidence is grounded during response generation~\cite{llavadi, compodistill, vgs}, even though output tokens are fundamentally a byproduct of the model attending to visual inputs.

\begin{figure}[t]
    \centering
    \includegraphics[width=1.0\linewidth]{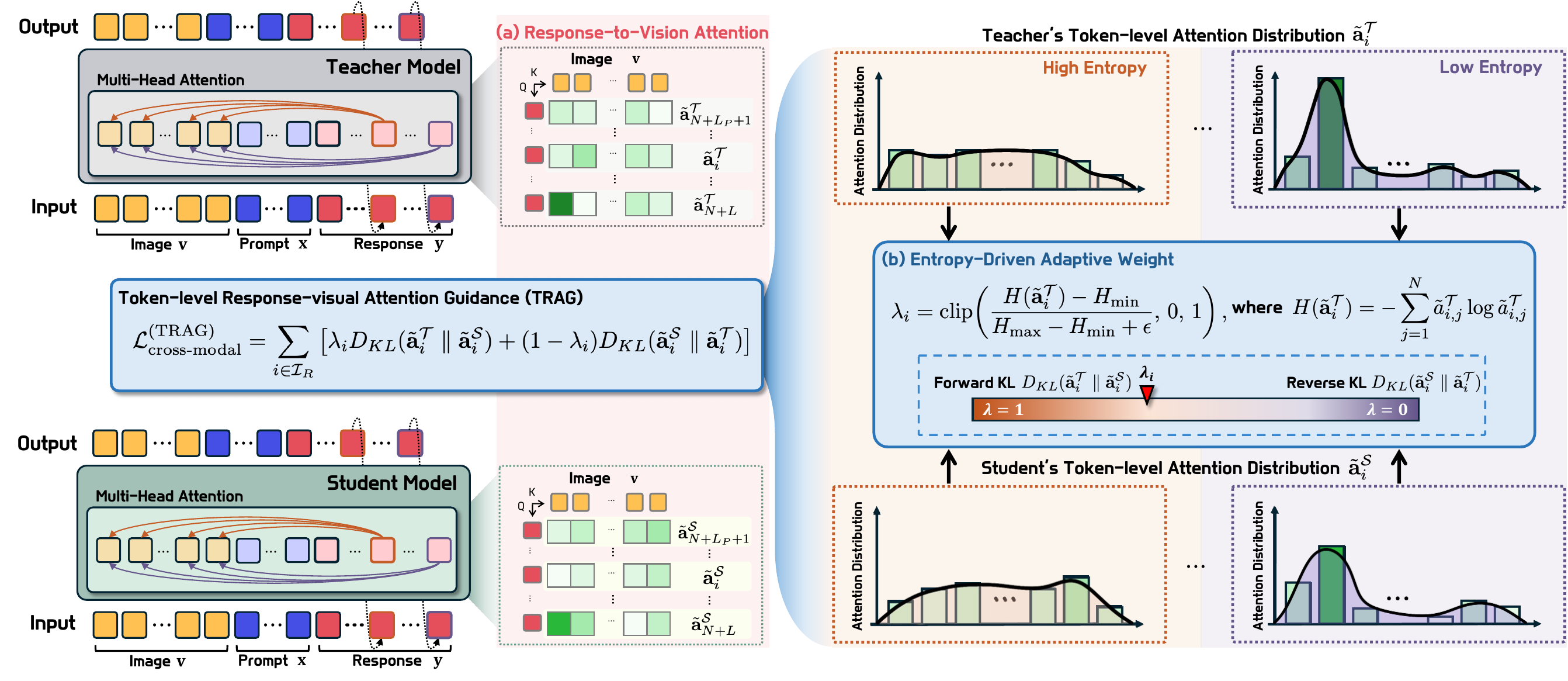}
    \caption{\textbf{Overview of Token-level Response-visual Attention Guidance (TRAG).} (a) TRAG distills cross-modal grounding by isolating Response-to-Vision attention, anchoring the student's attention distribution ($\tilde{\mathbf{a}}^{\mathcal{S}}_{i}$) to the teacher's ($\tilde{\mathbf{a}}^{\mathcal{T}}_{i}$). (b) It dynamically modulates the distillation objective $\mathcal{L}_{\text{cross-modal}}^{(\text{\method})}$ for each response token via an entropy-driven adaptive weight $\lambda_i$, derived from the teacher's attention distribution entropy $H(\tilde{\mathbf{a}}^{\mathcal{T}}_{i})$. By shifting this weight toward the mean-seeking Forward KL for high-entropy distributions and the mode-seeking Reverse KL for low-entropy targets, TRAG ensures the student accurately mirrors the teacher’s dynamic visual grounding during response generation.}
    \label{fig:overview}
    \vspace{-10pt}
\end{figure}

To provide a more direct supervisory signal, prior works have explored explicitly distilling cross-modal attention~\cite{alignkd, compodistill}. While this explicit guidance shows promise, identifying which attention signals to distill and at what granularity to apply supervision remains critically under-explored. Consequently, we find that the design choices made in existing frameworks often yield suboptimal supervision, failing to capture the dynamic evidence allocation that occurs during response decoding. In this work, we systematically characterize cross-modal attention distillation and its failure modes under existing design choices. 

\textbf{In detail, our contributions can be summarized as follows:}

\begin{itemize}[left=0em]
    \vspace{-5pt}
    \item[$\bullet$] \textbf{Response-to-Vision Attention as the Critical Signal:} We show that downstream performance correlates strongly with the alignment of Response-to-Vision attention between the student and the teacher model, whereas matching Prompt-to-Vision attention, which has been the primary focus of prior works, provides negligible utility. This identifies the generative phase, rather than the initial perception phase, as the critical signal for multimodal distillation.
    \item[$\bullet$] \textbf{Exposing the Limitations of Uniform Objectives:} We observe that cross-modal attention patterns exhibit significant variance across individual response tokens depending on their specific grounding roles, ranging from diffuse, scene-level context for function words to sharp, localized focus for visual entities. This high variance indicates that applying a uniform distillation objective across all tokens is suboptimal, necessitating an adaptive, token-wise supervision framework.
    \item[$\bullet$] Motivated by these findings, we propose \textbf{Token-level Response-visual Attention Guidance (TRAG)} (see Figure \ref{fig:overview}). TRAG shifts the distillation focus entirely to Response-to-Vision signals and introduces an adaptive, token-specific objective. By modulating the Kullback-Leibler (KL) divergence based on the teacher attention entropy, TRAG balances coverage and precision for each decoding step. Experiments on diverse multimodal benchmarks demonstrate that TRAG significantly outperforms competitive distillation baselines, achieving superior attention fidelity and robust performance gains.
\end{itemize}

\vspace{-20pt}

\section{Related Work}
\subsection{Knowledge Distillation in MLLMs}
Prior KD approaches for MLLMs largely follow two trajectories. One focuses on improving modality-specific representation alignment and architectural adaptations. LLaVADI~\cite{llavadi} analyzes key components for MLLM distillation, while LLaVA-KD~\cite{llavakd} extends logit-based distillation to the vision modality and enforces structural consistency among visual tokens. Related compression efforts further improve compact MLLMs through visual-encoder mixtures~\cite{movekd} or MoE-based architectures~\cite{llavamod}. While effective, these approaches do not explicitly supervise how visual evidence is selected during response generation, leaving visual grounding largely implicit.

Seeking more direct guidance on cross-modal processing, another trajectory explores attention-based distillation. Align-KD~\cite{alignkd} transfers Prompt-to-Vision attention at a selected decoder layer, while CompoDistill~\cite{compodistill} supervises Prompt-to-Vision attention maps. Concurrently, Align-TI~\cite{alignti} studies token-interaction distillation for multimodal alignment. However, these objectives mainly emphasize prompt-conditioned or globally defined interaction signals, overlooking token-wise dynamics during response generation. This gap motivates attention distillation that better aligns with response-time cross-modal processing, harmonizing with language modeling and logit distillation objectives.

\vspace{-10pt}

\subsection{Attention as a Distillation Signal}
Attention has evolved into a key supervisory signal for knowledge distillation, serving as a medium to transfer both spatial focus and complex inductive biases. Early methodologies focused on establishing correspondence between attention maps or saliency patterns to guide the student’s spatial perception toward task-relevant visual cues~\cite{at, catkd}. Subsequent developments introduced more structured forms of transfer, such as cross-modal structural constraints and sparse connectivity to highlight salient relationships~\cite{cacr, sakd}. More recently, attention supervision has been integrated with representation transfer to better preserve the teacher’s intermediate reasoning pathways and feature hierarchies~\cite{aftkd}. These advancements establish attention as a robust mechanism for capturing the teacher's internal prioritization, providing the technical groundwork for specialized attention-based strategies in the multimodal domain.

\begin{wrapfigure}{r}{0.4\textwidth}
    \centering
    \vspace{-30pt}
    \includegraphics[width=1.0\linewidth]{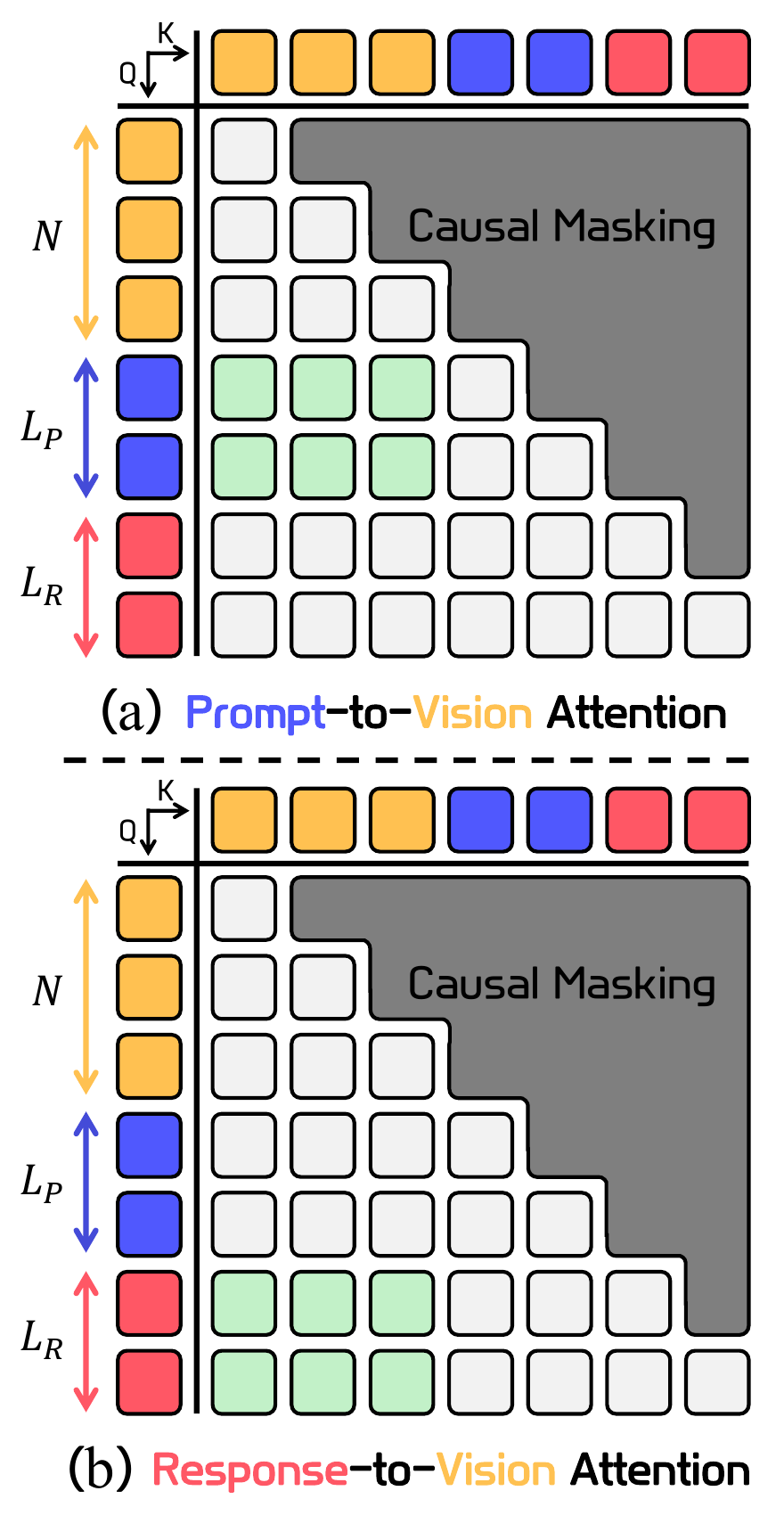}
    \vspace{-20pt}
    \caption{Comparison of attention extraction according to query source.}
    \label{fig:p2v_r2v_attention}
    \vspace{-10pt}
\end{wrapfigure}

\vspace{-10pt}

\section{Preliminaries} \label{sec:pre}

\vspace{-10pt}

\subsection{Cross-modal Attention Extraction} \label{sec:pre_attn}
Standard MLLMs encode an input image into visual tokens $\mathbf{v}=\{v_1,\dots,v_N\}$ via a vision encoder and an MLP-based projection module~\cite{llava}. The model then processes the concatenated sequence $\mathbf{c} = [\mathbf{v};\mathbf{x};\mathbf{y}]$, where $\mathbf{x}=\{x_1,\dots,x_{L_P}\}$ denotes the prompt token sequence and  $\mathbf{y}=\{y_1,\dots,y_{L_R}\}$ denotes the autoregressive response token sequence. Letting $L=L_P+L_R$, we index the elements $c_i$ of $\mathbf{c}$ such that indices $i\in \{1,...,N\}$, $i\in \{N+1,...,N+L_P\}$, and $i \in \{N+L_P+1,...,N+L\}$ map to the visual, prompt, and response tokens, respectively.

The generative process is driven by a Transformer decoder, whose self-attention at each layer $l\in\{1,\dots,M\}$ produces an attention matrix $\mathbf{A}_l\in\mathbb{R}^{(N+L)\times(N+L)}$ with entries $a_{l,i,j}$~\cite{transformers}. Under causal masking, each query index $i$ can attend only to keys $j\le i$. In particular, any textual query $i\in\{N+1,\dots,N+L\}$ can attend to visual keys $j\in\{1,\dots,N\}$ through cross-modal attention.

For any index set $\mathcal{I}\subseteq\{N+1,\dots,N+L\}$ of textual queries, we define the cross-modal attention submatrix from $\mathcal{I}$ to visual keys as:
\begin{equation}
\mathbf{A}_l(\mathcal{I}) := \mathbf{A}_l[\mathcal{I},\, 1\!:\!N] \in \mathbb{R}^{|\mathcal{I}|\times N},
\label{eq:submatrix_def}
\end{equation}
where $\mathbf{A}_l(\mathcal{I})[i,j]=a_{l,i,j}$ for $i\in\mathcal{I}$ and $j\in\{1,\dots,N\}$. In this paper, we consider two instantiations of $\mathcal{I}$: Prompt-to-Vision attention uses $\mathcal{I}_P=\{N+1,\dots,N+L_P\}$, whereas Response-to-Vision attention uses $\mathcal{I}_R=\{N+L_P+1,\dots,N+L\}$.
\begin{itemize}[left=0em]
    \item[$\bullet$] \textbf{Prompt-to-Vision Attention (}$\mathbf{A}_{P \to V}$, Figure \ref{fig:p2v_r2v_attention}-(a)\textbf{):} For each prompt token in $\mathbf{x}$ (i.e., $c_i$, where $i \in \mathcal{I}_P$), the distribution $\{a_{l,i,j}\}_{j=1}^{N}$ captures its attention over visual tokens during instruction encoding. This represents how strongly the model integrates visual evidence when interpreting the prompt token.
    \item[$\bullet$] \textbf{Response-to-Vision Attention (}$\mathbf{A}_{R \to V}$, Figure \ref{fig:p2v_r2v_attention}-(b)\textbf{):} For each response token in $\mathbf{y}$ (i.e., $c_i$, where $i \in \mathcal{I}_R$), the distribution $\{a_{l,i,j}\}_{j=1}^{N}$ captures which visual tokens the model attends to when generating the next response token. This represents a token-level trace of how visual context influences the response distribution at position $i$.
\end{itemize}

\vspace{-10pt}

\subsection{Knowledge Distillation Objectives for MLLMs} \label{sec:pre_mllmkd}
In prior works, the distillation objective for transferring knowledge from a teacher model $\mathcal{T}$ to a student model $\mathcal{S}$ is defined as a composite loss consisting of language modeling, uni-modal distillation, and cross-modal attention distillation:

\begin{equation}
\mathcal{L} = \mathcal{L}_{\text{LM}} + \mathcal{L}_{\text{uni-modal}} + \mathcal{L}_{\text{cross-modal}}.
\label{eq:L_total}
\end{equation}

\vspace{-15pt}

\subsubsection{Language Modeling Loss ($\mathcal{L}_{\text{LM}}$).}
The language modeling loss is the token-level cross entropy (negative log-likelihood) of the target response $\mathbf{y}$ conditioned on $\mathbf{v}$ and $\mathbf{x}$:
\begin{equation}
\mathcal{L}_{\text{LM}} = -\sum_{t=1}^{L_R} \log p^{\mathcal{S}}(y_t \mid \mathbf{v}, \mathbf{x}, y_{<t}).
\label{eq:L_LM}
\end{equation}

\vspace{-15pt}

\subsubsection{Uni-modal Knowledge Distillation Loss ($\mathcal{L}_{\text{uni-modal}}$).}
Uni-modal distillation aligns the teacher and student within each modality. 
This objective applies logit distillation for tokens in the response, alongside an auxiliary loss $\mathcal{L}_{\text{aux}}$:
\begin{equation}
\mathcal{L}_{\text{uni-modal}} = \sum_{t=1}^{L_R} D_{KL} \left( p^{\mathcal{T}}(\cdot \mid \mathbf{v}, \mathbf{x}, y_{<t}) \parallel p^{\mathcal{S}}(\cdot \mid \mathbf{v}, \mathbf{x}, y_{<t}) \right) + \mathcal{L}_{\text{aux}}.
\label{eq:L_uni}
\end{equation}
The first term performs response-level logit distillation over the target response sequence $\mathbf{y}=\{y_t\}_{t=1}^{L_R}$ \cite{hintonkd} by minimizing the KL divergence $D_{KL}(\cdot \parallel \cdot)$ between the teacher and student next-token distributions at each generation step.
The auxiliary loss $\mathcal{L}_{\text{aux}}$ introduces additional uni-modal supervision, typically defined on visual-side outputs or intra-visual relations. In LLaVA-KD \cite{llavakd}, $\mathcal{L}_{\text{aux}}$ distills vision-token logits and matches affinity matrices among vision tokens. Similarly, CompoDistill \cite{compodistill} distills intermediate-layer visual self-attention by aligning vision-to-vision attention maps with a cosine-based loss.\footnote{We decompose the single attention distillation loss in CompoDistill \cite{compodistill} into uni-modal and cross-modal terms. The detailed derivation is provided in Appendix~\ref{appendix:compo_formulation}.}
Overall, $\mathcal{L}_{\text{aux}}$ strengthens modality-specific alignment by providing additional supervision beyond response-level logit distillation.

\vspace{-10pt}

\subsubsection{Cross-modal Knowledge Distillation Loss ($\mathcal{L}_{\text{cross-modal}}$).} 
Cross-modal distillation aligns the teacher and student by matching their cross-modal attention maps, where textual queries attend to visual tokens. In CompoDistill \cite{compodistill}, this objective is instantiated with a group-based layer matching scheme to handle depth differences between the teacher and student:

\begin{equation}
\mathcal{L}_{\text{cross-modal}}^{\text{(CompoDistill)}}
= \sum_{l \in \mathcal{M}_\mathcal S}
\mathcal{D}_{\text{cos}}\!\left(
\mathbf{A}^{\mathcal{T}}_{G_l}(\mathcal{I}),\ 
\mathbf{A}^{\mathcal{S}}_{l}(\mathcal{I})
\right),
\label{eq:L_cross}
\end{equation}
where $\mathcal{D}_{\text{cos}}(\cdot,\cdot)$ denotes cosine distance, $\mathcal{M}_{\mathcal S}$ denotes a subset of student layers selected for supervision, and $\mathcal{I}$ denotes the textual query indices used to extract $\mathbf{A}_l(\mathcal{I})$ (Eq.~\ref{eq:submatrix_def}). For each student layer $l$, where $G_l$ denotes the set of teacher layers assigned to $l$, the grouped teacher attention is
\begin{equation}
\mathbf{A}^{\mathcal{T}}_{G_l}(\mathcal{I})
:= \frac{1}{|G_l|}\sum_{g\in G_l} \mathbf{A}^{\mathcal{T}}_{g}(\mathcal{I}).
\label{eq:teacher_group_avg}
\end{equation}
In prior work, including Align-KD~\cite{alignkd} and CompoDistill~\cite{compodistill}, \textbf{$\mathcal{I}$ is typically chosen as the prompt index range $\{N+1,\dots,N+L_P\}$} (i.e., they set $\mathcal{I}$ to $\mathcal{I}_P$), focusing supervision on Prompt-to-Vision attention. 

\vspace{-5pt}

\section{Revisiting Cross-modal Knowledge Distillation} \label{sec:obs}
\vspace{-5pt}
Building on Section~\ref{sec:pre}, we revisit two common but under-discussed choices in attention-based cross-modal knowledge distillation. First, prior work often supervises attention on prompt tokens by default. Second, they typically devote substantial design effort to deciding which layers to distill from and how to align layers across teachers and students with different depths. \textit{In this section, we ask whether prompt-based supervision is truly optimal, and whether layer selection is as critical as often assumed compared to the granularity of the attention signal.}

\begin{figure}[b]
    \vspace{-5pt}
    \centering
    \includegraphics[width=1\linewidth]{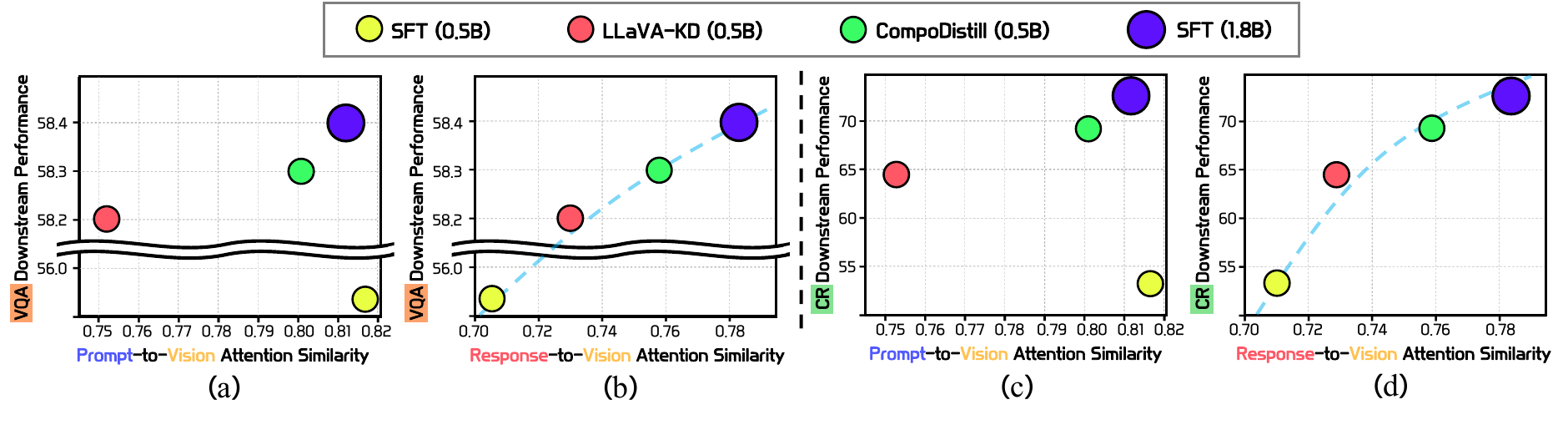}
    \captionsetup{skip=0.5pt}
    \caption{\textbf{Correlation analysis between attention similarity and downstream performance.} (a) and (c) show that Prompt-to-Vision attention similarity exhibits negligible correlation with downstream performance on Visual Question Answering and Compositional Reasoning tasks. In contrast, (b) and (d) demonstrate a strong positive correlation between Response-to-Vision attention similarity and model performance, identifying the generative phase as the critical signal for multimodal distillation.}
    \label{fig:obs1}
    \vspace{-20pt}
\end{figure}

\vspace{-5pt}

\subsection{Is Aligning Prompt-to-Vision Attention to the Teacher Optimal for Knowledge Transfer?} \label{sec:obs_response}
If Prompt-to-Vision attention alignment to the teacher is an optimal distillation target, then Prompt-to-Vision attention similarity should consistently accompany a stronger student. To examine this, we compute teacher-student attention similarity and analyze its correlation with benchmark performance across multiple students.
We measure teacher–student cross-modal attention similarity as cosine similarity under teacher forcing on an unseen dataset \textit{VIGC-InstData}~\cite{vigc}.

The results show that prompt-to-vision alignment does not reliably track model capability. As shown in Figures~\ref{fig:obs1}-(a) and (c), in Prompt-to-Vision attention ($\mathbf{A}_{P\to V}$), similarity to the teacher does not consistently correlate with downstream performance. Notably, the Supervised Fine-Tuning (SFT) baseline can achieve higher $\mathbf{A}_{P\to V}$ similarity than distilled models such as LLaVA-KD and CompoDistill, while still underperforming them on downstream benchmarks. This indicates that aligning attention during prompt processing alone is not a sufficient proxy for the cross-modal behavior that distinguishes stronger models.

In contrast, Response-to-Vision alignment is substantially more predictive. Figures~\ref{fig:obs1}-(b) and (d) show a strong positive correlation between Response-to-Vision attention ($\mathbf{A}_{R\to V}$) similarity and performance across students. This suggests that the teacher’s effective visual grounding is expressed more clearly during response generation than during prompt processing, motivating attention supervision that targets the generative phase.

\begin{figure}[h]
    \vspace{-20pt}
    \centering
    \includegraphics[width=1.0\linewidth]{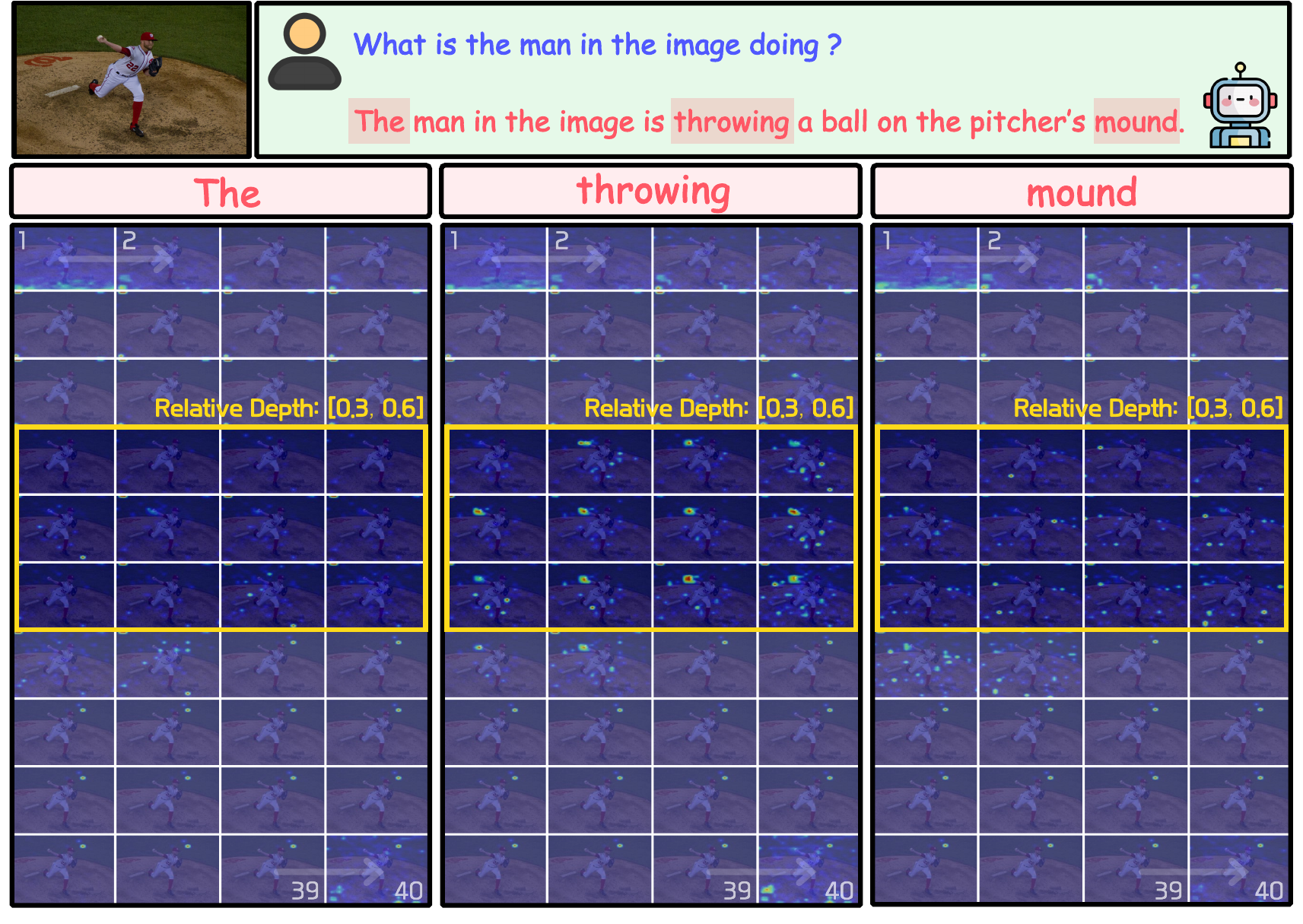}
    \captionsetup{skip=0.5pt}
    \caption{\textbf{Analysis of attention granularity across layers and tokens.} Visualization of teacher attention maps reveals that while attention patterns exhibit high redundancy across adjacent layers, they show significant variance across individual response tokens. The yellow box marks intermediate layers (relative depth in $[0.3,0.6]$ on a $[0,1]$ normalized depth axis), where cross-modal interactions are most pronounced. The model exhibits weak, diffuse attention when generating the function word \textit{`The'}, whereas it sharply concentrates on the pitcher's hand when generating \textit{`throwing'} and shifts to the surrounding ground when generating \textit{`mound'}.}
    \label{fig:obs2}
\end{figure}

\subsection{Which Granularity Governs Knowledge Transfer: Layers or Tokens?} \label{sec:obs_pertoken}
Section~\ref{sec:obs_response} suggests that the key transferable signal emerges during generation, where Response-to-Vision attention ($\mathbf{A}_{R\to V}$) tracks downstream performance. The next question is the granularity at which this signal should be distilled. Most attention distillation methods adopt layer-level granularity through layer selection or mapping. We ask whether this layer-level choice is necessary by analyzing $\mathbf{A}_{R\to V}$ across both decoder depth and individual response tokens.

Focusing on intermediate depths (relative depth in $[0.3,0.6]$), where cross-modal interactions are most pronounced~\cite{compodistill, focus_area, intermediate_layer, inter1, inter2}, Figure~\ref{fig:obs2} shows that adjacent layers exhibit highly similar $\mathbf{A}_{R\to V}$ patterns, indicating substantial redundancy along depth. In contrast, $\mathbf{A}_{R\to V}$ varies sharply across response tokens. When generating function words (e.g., articles), the model exhibits weak visual grounding with uniform, blurry attention, whereas content tokens show more diverse patterns ranging from sharply localized evidence to diffuse, scene-level grounding. This suggests that the variation in cross-modal grounding is primarily token-dependent rather than layer-dependent, motivating token-wise supervision that adapts to each response token. Additional examples are provided in Appendix~\ref{appendix:obs2}.

\section{Token-level Response-visual Attention Guidance} \label{sec:method}
Motivated by our findings that Response-to-Vision attention alignment is a strong predictor of downstream task proficiency, and that cross-modal attention patterns vary substantially across response tokens (Section~\ref{sec:obs}), we propose \textbf{\underline{T}oken-level \underline{R}esponse-visual \underline{A}ttention \underline{G}uidance~(\method)}. 

Building on the standard training objective (Eq.~\ref{eq:L_total}), we take the CompoDistill formulation (Eq.~\ref{eq:L_cross}) as a starting point for the cross-modal distillation term, $\mathcal{L}_{\text{cross-modal}}$. \method~reformulates this baseline into a response-token-wise objective that guides the student's Response-to-Vision attention using the teacher's attention as a reference for each target response token.

\subsection{Shifting Supervision to the Response Phase} \label{sec:method_response}
Motivated by the observation in Section~\ref{sec:obs_response}, we shift cross-modal attention supervision from the prompt to the response, where visual grounding is expressed during generation. Accordingly, we replace the prompt index set $\mathcal{I}_P$ with the response range $\mathcal{I}_R=\{N+L_P\!+\!1,\dots,N+L\}$:
\begin{equation}
\mathcal{L}_{\text{cross-modal}}^{\text{(response)}}
= \sum_{l\in\mathcal M_\mathcal S}\mathcal D_{\cos}\!\left(\mathbf A^{\mathcal T}_{G_l}(\mathcal{I}_R),\mathbf A^{\mathcal S}_{l}(\mathcal{I}_R)\right).
\label{eq:L_response}
\end{equation} 
This localizes attention supervision to response generation directly. Next, we refine the granularity of the distillation signal within this response range.

\subsection{Adaptive Granularity via Entropy-Aware Weighting} \label{sec:method_entropy_wkl}
\subsubsection{Shifting from Global to Token-Level Supervision.}

Prior attention-based distillation methods typically define a single uniform objective over the supervised query set, comparing the teacher and student attention maps using a fixed discrepancy measure. However, Section~\ref{sec:obs_pertoken} shows that token-level attention distributions are highly heterogeneous, suggesting that a single uniform constraint provides limited resolution to preserve per-token distributional intent. Moreover, Section~\ref{sec:obs_pertoken} indicates substantial redundancy across intermediate layers, motivating aggregation across layers rather than sensitive layer selection.

Accordingly, we first aggregate cross-modal attention over intermediate layers to obtain a representative attention distribution for each response query, and then decompose supervision into token-level losses. Let $\mathcal{M}_ \mathcal S$ and $\mathcal{M}_\mathcal T$ denote the sets of intermediate layers (relative depth in $[0.3,0.6]$) used for aggregation in the student and teacher, respectively. For each response query index $i\in\mathcal{I}_R$, we denote the per-layer attention vectors over $N$ visual tokens as
$\mathbf{a}^{\mathcal{S}}_{l,i}:=\{a^{\mathcal{S}}_{l,i,j}\}_{j=1}^{N}$ and
$\mathbf{a}^{\mathcal{T}}_{l,i}:=\{a^{\mathcal{T}}_{l,i,j}\}_{j=1}^{N}$,
corresponding to the rows of $\mathbf{A}^{\mathcal{S}}_{l}(\mathcal{I}_R)$ and $\mathbf{A}^{\mathcal{T}}_{l}(\mathcal{I}_R)$ restricted to visual keys (Eq.~\ref{eq:submatrix_def}). 
We then obtain layer-averaged attention vectors by averaging over intermediate layers $\mathcal{M}_ \mathcal S$ and $\mathcal{M}_ \mathcal T$:
\begin{equation}
\bar{\mathbf{a}}^{\mathcal{S}}_{i}
:= \frac{1}{|\mathcal{M}_\mathcal S|}\sum_{l\in\mathcal{M}_\mathcal S} \mathbf{a}^{\mathcal{S}}_{l,i},
\qquad
\bar{\mathbf{a}}^{\mathcal{T}}_{i}
:= \frac{1}{|\mathcal{M}_\mathcal T|}\sum_{l\in\mathcal{M}_\mathcal T} \mathbf{a}^{\mathcal{T}}_{l,i}.
\label{eq:layer_agg_ai}
\end{equation}

Since these vectors are obtained by restricting attention to visual keys, their total mass over $\{1,\dots,N\}$ is not necessarily one. 
We therefore renormalize them on the vision span to form proper distributions:
\begin{equation}
\tilde{\mathbf{a}}^{\mathcal{S}}_{i}
:= \frac{\bar{\mathbf{a}}^{\mathcal{S}}_{i}}{\sum_{j=1}^{N}\bar{a}^{\mathcal{S}}_{i,j}+\epsilon},
\qquad
\tilde{\mathbf{a}}^{\mathcal{T}}_{i}
:= \frac{\bar{\mathbf{a}}^{\mathcal{T}}_{i}}{\sum_{j=1}^{N}\bar{a}^{\mathcal{T}}_{i,j}+\epsilon},
\label{eq:vision_renorm}
\end{equation}
where $\epsilon>0$ is a small constant for numerical stability. 
Finally, we define the token-level cross-modal objective as
\begin{equation}
\mathcal{L}_{\text{cross-modal}}^{\text{(token-level)}}
= \sum_{i\in\mathcal{I}_R}
\mathcal{D}_{\text{guidance}}\!\left(
\tilde{\mathbf{a}}^{\mathcal{T}}_{i},\ 
\tilde{\mathbf{a}}^{\mathcal{S}}_{i}
\right),
\label{eq:L_tokenlevel}
\end{equation}
where $\mathcal{D}_{\text{guidance}}(\cdot,\cdot)$ is defined in the following subsection in detail.

\subsubsection{Entropy-Driven Adaptive Weighting.}
Having decomposed the cross-modal objective into per-token terms (Eq.~\ref{eq:L_tokenlevel}), we now derive a token-dependent coefficient that modulates the guidance behavior. For each response query index $i\in\mathcal{I}_R$, we compute the entropy of the teacher’s normalized layer-aggregated attention distribution $\tilde{\mathbf{a}}_{i}^{\mathcal{T}}$ (Eq.~\ref{eq:vision_renorm}):

\begin{equation}
H(\tilde{\mathbf{a}}_{i}^{\mathcal{T}})
= -\sum_{j=1}^{N}\tilde{a}^{\mathcal{T}}_{i,j}\log \tilde{a}^{\mathcal{T}}_{i,j},
\label{eq:entropy_calculation}
\end{equation}
where $\tilde{a}^{\mathcal{T}}_{i,j}$ denotes the weight on the $j$-th visual token ($j\in\{1,\dots,N\}$) in $\tilde{\mathbf{a}}_{i}^{\mathcal{T}}$.

We map $H(\tilde{\mathbf{a}}_{i}^{\mathcal{T}})$ to a bounded coefficient $\lambda_i\in[0,1]$ via min--max normalization with clipping:
\begin{equation}
\lambda_i
= \mathrm{clip}\!\left(
\frac{H(\tilde{\mathbf{a}}_{i}^{\mathcal{T}})-H_{\min}}{H_{\max}-H_{\min}+\epsilon},
\,0,\,1
\right),
\label{eq:lambda}
\end{equation}
where $H_{\min}$ and $H_{\max}$ are running lower and upper bounds updated online as exponential moving averages (momentum $\beta=0.995$) of the mini-batch minimum and maximum of $H(\tilde{\mathbf{a}}_{i}^{\mathcal{T}})$, and $\epsilon>0$ is a small constant for numerical stability. A larger $\lambda_i$ corresponds to more diffuse attention, which we leverage to parameterize token-wise guidance in the following objective.

\subsubsection{Distributional Guidance via Adaptive KL Regulation.}
To achieve specialized supervision, we define the guidance operator $\mathcal{D}_{\text{guidance}}$ as a weighted combination of forward and reverse Kullback-Leibler (KL) divergences. This formulation leverages the asymmetry of KL divergence. Forward KL, $D_{KL}(\tilde{\mathbf{a}}_{i}^{\mathcal{T}}\parallel\tilde{\mathbf{a}}_{i}^{\mathcal{S}})$, strongly penalizes the student when it assigns low probability to regions where the teacher has non-negligible mass, thereby promoting coverage. In contrast, reverse KL, $D_{KL}(\tilde{\mathbf{a}}_{i}^{\mathcal{S}}\parallel\tilde{\mathbf{a}}_{i}^{\mathcal{T}})$, penalizes probability mass placed outside the teacher’s high-density regions more aggressively, which encourages concentration on dominant modes \cite{bdkd, akl}.

Using the normalized coefficient $\lambda_i$ defined in Eq. \ref{eq:lambda}, we combine the two asymmetric KL terms into a token-wise guidance divergence per response token:
\begin{equation}
\mathcal{D}_{\text{guidance}}(\tilde{\mathbf{a}}_{i}^{\mathcal{T}}, \tilde{\mathbf{a}}_{i}^{\mathcal{S}}) = \lambda_i D_{KL}(\tilde{\mathbf{a}}_{i}^{\mathcal{T}} \parallel \tilde{\mathbf{a}}_{i}^{\mathcal{S}}) + (1 - \lambda_i) D_{KL}(\tilde{\mathbf{a}}_{i}^{\mathcal{S}} \parallel \tilde{\mathbf{a}}_{i}^{\mathcal{T}}).
\label{eq:L_wkl}
\end{equation}
Substituting Eq.~\ref{eq:L_wkl} into Eq.~\ref{eq:L_tokenlevel} yields the final cross-modal objective (Figure \ref{fig:overview}):
\begin{equation}
\mathcal{L}_{\text{cross-modal}}^{(\text{\method})}
= \sum_{i\in\mathcal{I}_R}
\left[
\lambda_i\, D_{KL}(\tilde{\mathbf{a}}_{i}^{\mathcal{T}} \parallel \tilde{\mathbf{a}}_{i}^{\mathcal{S}})
+ (1-\lambda_i)\, D_{KL}(\tilde{\mathbf{a}}_{i}^{\mathcal{S}} \parallel \tilde{\mathbf{a}}_{i}^{\mathcal{T}})
\right].
\label{eq:trag_cross_modal_final}
\end{equation}
This adaptive regulation overcomes the resolution limits of uniform objectives, enabling the student to capture visual evidence from localized to diffuse patterns.

\subsection{Overall Training Framework}
\label{sec:overall_training}

We adopt a three-stage training schedule (DPT--SFT--DFT), consistent with prior MLLM distillation works~\cite{llavakd,compodistill}. DPT trains the projection module on image--caption data with the language backbone frozen, SFT improves instruction following with the standard language modeling objective, and DFT re-applies teacher-guided distillation on the instruction data. The objective is stage-dependent: in DPT and DFT, we optimize
\begin{equation}
\mathcal{L}_{\text{\method}}
= \mathcal{L}_{\text{LM}} + \mathcal{L}_{\text{uni-modal}} + \mathcal{L}_{\text{cross-modal}}^{(\text{\method})},
\label{eq:trag_loss}
\end{equation}
whereas SFT minimizes only $\mathcal{L}_{\text{LM}}$. In Eq.~\ref{eq:trag_loss}, $\mathcal{L}_{\text{uni-modal}}$ follows the uni-modal distillation components of LLaVA-KD~\cite{llavakd}, and $\mathcal{L}_{\text{cross-modal}}^{(\text{\method})}$ is defined in Eq.~\ref{eq:trag_cross_modal_final}. Detailed formulations are provided in Appendix~\ref{appendix:loss_detail}.

\section{Experiments}
\subsection{Experimental Setup}
\subsubsection{Implementation Details.}
Both the teacher and student models use the same vision encoder, SigLIP-SO400M/14 at 384px resolution~\cite{siglip}, and their LLM backbones are drawn from the Qwen1.5 and Qwen2.5 families. A two-layer MLP with GELU nonlinearity serves as the projection module to map visual features into the language embedding space \cite{gelu}.

We use \textit{LLaVA-Pretrain-558K} for the DPT stage and \textit{LLaVA-Instruct-665K} for the SFT and DFT stages~\cite{llava1.5}. Throughout all stages, the teacher model is frozen. For the student, we train only the projection module in the DPT stage, keeping the vision encoder and LLM frozen. In the SFT and DFT stages, we freeze the vision encoder and fully finetune the projection module and the LLM. All models are trained on 2 NVIDIA A100 GPUs with DeepSpeed ZeRO-2. Detailed hyperparameters are provided in the Appendix~\ref{appendix:implementation}.

\subsubsection{Evaluation Benchmarks.}
We evaluate our models on two benchmark suites covering general Visual Question Answering (VQA) and compositional reasoning (CR). The general VQA suite consists of eight benchmarks, including GQA~\cite{gqa}, ScienceQA~\cite{sqa}, TextVQA~\cite{textvqa}, MME~\cite{mme}, MMBench in English and Chinese~\cite{mmb}, POPE~\cite{pope}, and MMMU~\cite{mmmu}. We further evaluate on three CR benchmarks: SugarCrepe~\cite{sugarcrepe}, BiVLC~\cite{bivlc}, and Winoground~\cite{winoground}. Detailed evaluation metrics and setup are described in the Appendix~\ref{appendix:evaluation}.

\begin{table*}[t]
\centering
\caption{
\textbf{General VQA benchmark comparison of \method.} Performance on eight general VQA benchmarks across multiple backbones and parameter scales. Models are grouped by size; \# Samples denotes the number of training samples.  $Avg_8$ denotes the average over all 8 benchmarks, while $Avg_6$ averages the remaining 6 benchmarks excluding MMB$^{\text{CN}}$ and MMMU. For the $\leq$ 2B and $\leq$ 0.5B categories, the best performing result is highlighted in \textbf{bold} and the second-best result is indicated with \underline{underline}. \sethlcolor{blue}\hl{Blue} indicates the teacher model for each backbone, \sethlcolor{gray}\hl{gray} rows represent models trained with the standard PT-SFT pipeline, \sethlcolor{yellow}\hl{yellow} denotes recent competitive distillation baselines, and \sethlcolor{green}\hl{green} indicates our proposed method. The symbol $\dag$ denotes results reproduced under our experimental setup for fair comparison.
}
\vspace{-5pt}
\renewcommand{\arraystretch}{0.9}
\resizebox{1.\linewidth}{!}{
\begin{tabular}{c!{\vrule width 0.8pt}ccc!{\vrule width 0.8pt}cccccccc!{\vrule width 0.8pt}cc}
\Xhline{1.5pt}
\multirow{2}{*}{Size} & \multirow{2}{*}{Method} & \multirow{2}{*}{LLM} & \multirow{2}{*}{\# Samples} & \multicolumn{8}{c|}{Visual Question Answering Benchmarks} & \multirow{2}{*}{$Avg_6$} & \multirow{2}{*}{$Avg_8$} \\\Xcline{5-12}{1pt}
                                 &                &       &                            & GQA   & ScienceQA  & TextVQA  & MME  & MMB  & $\text{MMB}^{\text{CN}}$ & POPE & MMMU & &  \\\Xhline{1pt}
\multirow{9}{*}{$\leq$ 4B} 
& Bunny                            & Phi2-2.7B            & 2.6 M                      & 62.5  & 70.9   & 56.7     & 74.4 & 68.6 &        37.2              & -    & 38.2 & - & -                   \\
& Imp-3B                           & Phi2-2.7B            & 1.5 M                      & 63.5  & 72.8   & 59.8     & -    & 72.9 &        46.7              & -    & -    & - & -                  \\
& MobileVLM                        & MobileLLaMA-2.7B          & 1.2 M                 & 59.0  & 61.0   & 47.5     & 64.4 & 59.6 &        -                 & 84.9 & -    & 62.7 & -                  \\
& MoE-LLaVA                        & Phi2-2.7B            & 2.2 M                      & 62.6  & 70.3   & 57.0     & -    & 68.0 &        -                 & 85.7 & -    & - & -                  \\
& MiniCPM-V                        & MiniCPM-2.4B         & 570 M                      & 51.5  & 74.4   & 56.6     & 68.9 & 64.0 &        62.7              & 79.5 & -    & 65.8 & -                  \\
& TinyLLaVA                        & Phi2-2.7B            & 1.2 M                      & 62.0  & 69.1   & 59.1     & 73.2 & 66.9 &        -                 & 86.4 & 38.4 & 69.4 & -                  \\
&\cellcolor{blue}\text{TinyLLaVA}$^\dag$          &\cellcolor{blue}Qwen1.5-4B           &\cellcolor{blue}1.2 M                         &\cellcolor{blue}62.7  &\cellcolor{blue}70.2   &\cellcolor{blue}60.0     &\cellcolor{blue}70.5 &\cellcolor{blue}67.5 &\cellcolor{blue}67.2              &\cellcolor{blue}86.9 &\cellcolor{blue}38.6 &\cellcolor{blue}69.6 &\cellcolor{blue}65.4               \\
&\cellcolor{blue}\text{TinyLLaVA}$^\dag$      &\cellcolor{blue}Qwen2.5-3B           &\cellcolor{blue}1.2 M                           &\cellcolor{blue}63.0     &\cellcolor{blue}76.1      &\cellcolor{blue}60.1        &\cellcolor{blue}71.6    &\cellcolor{blue}73.1    &\cellcolor{blue}70.4                 &\cellcolor{blue}87.8    &\cellcolor{blue}41.3    &\cellcolor{blue}71.9 &\cellcolor{blue}67.9                 \\
&\cellcolor{yellow}LLaVADI                          &\cellcolor{yellow}MobileLLaMA-2.7B          &\cellcolor{yellow}1.2 M                            &\cellcolor{yellow}61.4  &\cellcolor{yellow}64.1   &\cellcolor{yellow}50.7     &\cellcolor{yellow}68.8 &\cellcolor{yellow}62.5 &\cellcolor{yellow}-                 &\cellcolor{yellow}86.7 &\cellcolor{yellow}-    &\cellcolor{yellow}65.7 &\cellcolor{yellow}-                 \\\Xhline{1pt}
\multirow{15}{*}{$\leq$ 2B}
& Imp-2B                           & Qwen1.5-1.8B         & 1.5 M                      & 61.9  & 66.1   & 54.5     & 65.2 & 63.8 &        61.3              & 86.7 & -    & 66.3 & -               \\
& Bunny-2B                         & Qwen1.5-1.8B         & 2.6 M                      & 59.6  & 64.6   & 53.2     & 65.0 & 59.1 &        58.5              & 85.8 & -    & 64.5 & -               \\
& Mini-Gemini-2B                   & Gemma-2B             & 2.7 M                      & 60.7  & 63.1   & 56.2     & 67.0 & 59.8 &        51.3              & 85.6 & 31.7 & 65.4 & -               \\
& MoE-LLaVA-2B                     & Qwen1.5-1.8B         & 2.2 M                      & 61.5  & 63.1   & 48.0     & 64.6 & 59.7 &        57.3              & \underline{87.0} & -    & 63.9 & -               \\
&\cellcolor{gray}\text{TinyLLaVA}$^\dag$          &\cellcolor{gray}Qwen1.5-1.8B         &\cellcolor{gray}1.2 M                         &\cellcolor{gray}60.9  &\cellcolor{gray}65.2   &\cellcolor{gray}47.7     &\cellcolor{gray}61.3 &\cellcolor{gray}57.8 &\cellcolor{gray}56.2              &\cellcolor{gray}83.4 &\cellcolor{gray}34.8 &\cellcolor{gray}62.7 &\cellcolor{gray}58.4               \\
&\cellcolor{gray}\text{TinyLLaVA}$^\dag$          &\cellcolor{gray}Qwen2.5-1.5B         &\cellcolor{gray}1.2 M                            &\cellcolor{gray}61.7     &\cellcolor{gray}70.9      &\cellcolor{gray}58.4        &\cellcolor{gray}\underline{70.4}    &\cellcolor{gray}70.6    &\cellcolor{gray}64.0                 &\cellcolor{gray}86.0    &\cellcolor{gray}\textbf{39.2}    &\cellcolor{gray}69.7 &\cellcolor{gray}65.1                  \\
& \text{KD}$^\dag$                     & Qwen1.5-1.8B         & 1.2 M                       & 60.9  & 64.6   & 53.3     & 66.3 & 65.8 &        63.1              & 86.5 & 32.7    & 66.2 & 61.6              \\
&\cellcolor{yellow}LLaVADI                          &\cellcolor{yellow}MobileLLaMA-1.4B          &\cellcolor{yellow}1.2 M                      &\cellcolor{yellow}55.4  &\cellcolor{yellow}56.0   &\cellcolor{yellow}45.3     &\cellcolor{yellow}58.9 &\cellcolor{yellow}55.0 &\cellcolor{yellow}-                 &\cellcolor{yellow}84.7 &\cellcolor{yellow}-    &\cellcolor{yellow}59.2 &\cellcolor{yellow}-                   \\
&\cellcolor{yellow}LLaVA-MoD                          &\cellcolor{yellow}Qwen2-1.5B          &\cellcolor{yellow}5 M                            &\cellcolor{yellow}58.8  &\cellcolor{yellow}69.2   &\cellcolor{yellow}\underline{59.9}     &\cellcolor{yellow}69.2 &\cellcolor{yellow}68.9 &\cellcolor{yellow}64.4                 &\cellcolor{yellow}\textbf{87.2} &\cellcolor{yellow}-    &\cellcolor{yellow}68.8 &\cellcolor{yellow}-                  \\
&\cellcolor{yellow}Align-KD                          &\cellcolor{yellow}MobileLLaMA-1.4B          &\cellcolor{yellow}3.6 M                           &\cellcolor{yellow}60.1  &\cellcolor{yellow}67.7   &\cellcolor{yellow}53.1     &\cellcolor{yellow}65.1 &\cellcolor{yellow}57.5 &\cellcolor{yellow}-                 &\cellcolor{yellow}\underline{87.0} &\cellcolor{yellow}-    &\cellcolor{yellow}65.0 &\cellcolor{yellow}-     \\
&\cellcolor{yellow}\text{CompoDistill}$^\dag$                          &\cellcolor{yellow}Qwen1.5-1.8B          &\cellcolor{yellow}1.2 M                           &\cellcolor{yellow}61.2  &\cellcolor{yellow}66.5   &\cellcolor{yellow}53.5     &\cellcolor{yellow}67.0 &\cellcolor{yellow}64.5 &\cellcolor{yellow}63.0                 &\cellcolor{yellow}85.5 &\cellcolor{yellow}34.1    &\cellcolor{yellow}66.4 &\cellcolor{yellow}61.9     \\
&\cellcolor{yellow}LLaVA-KD                          &\cellcolor{yellow}Qwen1.5-1.8B          &\cellcolor{yellow}1.2 M                           &\cellcolor{yellow}62.3     &\cellcolor{yellow}64.7     &\cellcolor{yellow}53.4 &\cellcolor{yellow}69.1 &\cellcolor{yellow}64.0                 &\cellcolor{yellow}63.7 &\cellcolor{yellow}86.3    &\cellcolor{yellow}33.6 &\cellcolor{yellow}66.6 &\cellcolor{yellow}62.1                 \\
&\cellcolor{yellow}LLaVA-KD                          &\cellcolor{yellow}Qwen2.5-1.5B          &\cellcolor{yellow}1.2 M                          &\cellcolor{yellow}\underline{62.5}     &\cellcolor{yellow}\underline{71.6}     &\cellcolor{yellow}59.7 &\cellcolor{yellow}70.0 &\cellcolor{yellow}\textbf{71.0}                 &\cellcolor{yellow}\underline{66.6} &\cellcolor{yellow}86.7    &\cellcolor{yellow}35.8 &\cellcolor{yellow}\underline{70.2} &\cellcolor{yellow}\underline{65.4}    \\
&\cellcolor{green}\textbf{\method}                          &\cellcolor{green}Qwen1.5-1.8B         &\cellcolor{green}1.2 M                           &\cellcolor{green}61.7     &\cellcolor{green}66.9      &\cellcolor{green}55.6        &\cellcolor{green}67.0    &\cellcolor{green}66.6    &\cellcolor{green}63.3                 &\cellcolor{green}86.8    &\cellcolor{green}35.2    &\cellcolor{green}67.4 &\cellcolor{green}62.9                  \\ 
&\cellcolor{green}\textbf{\method}                          &\cellcolor{green}Qwen2.5-1.5B         &\cellcolor{green}1.2 M                          &\cellcolor{green}\textbf{62.6}     &\cellcolor{green}\textbf{73.2}      &\cellcolor{green}\textbf{60.0}        &\cellcolor{green}\textbf{71.2}    &\cellcolor{green}\underline{70.8}    &\cellcolor{green}\textbf{69.3}                 &\cellcolor{green}\textbf{87.2}    &\cellcolor{green}\underline{38.7}    &\cellcolor{green}\textbf{70.8} &\cellcolor{green}\textbf{66.6}                  \\\Xhline{1pt}
\multirow{9}{*}{$\leq$ 0.5B}
&\cellcolor{gray}\text{TinyLLaVA}$^\dag$          &\cellcolor{gray}Qwen1.5-0.5B         &\cellcolor{gray}1.2 M                         &\cellcolor{gray}56.7  &\cellcolor{gray}60.9   &\cellcolor{gray}46.6     &\cellcolor{gray}59.4 &\cellcolor{gray}55.2 &\cellcolor{gray}51.8              &\cellcolor{gray}83.9 &\cellcolor{gray}31.9 &\cellcolor{gray}60.4 &\cellcolor{gray}55.8               \\
&\cellcolor{gray}\text{TinyLLaVA}$^\dag$          &\cellcolor{gray}Qwen2.5-0.5B         &\cellcolor{gray}1.2 M                            &\cellcolor{gray}58.6     &\cellcolor{gray}61.2      &\cellcolor{gray}47.9        &\cellcolor{gray}62.3    &\cellcolor{gray}59.3    &\cellcolor{gray}52.4                 &\cellcolor{gray}85.1    &\cellcolor{gray}30.4    &\cellcolor{gray}62.4 &\cellcolor{gray}57.1               \\
& \text{KD}$^\dag$                     & Qwen1.5-0.5B         & 1.2 M                        & 56.8  & 60.4   & 46.1     & 58.9 & 54.7 &        49.7              & 84.9 & \underline{32.3}    & 60.3 &55.5               \\
&\cellcolor{yellow}LLaVA-MoD                        &\cellcolor{yellow}Qwen2-0.5B         &\cellcolor{yellow}5 M                              &\cellcolor{yellow}56.6  &\cellcolor{yellow}61.1   &\cellcolor{yellow}\textbf{57.1}     &\cellcolor{yellow}\underline{67.0} &\cellcolor{yellow}58.7 &\cellcolor{yellow}54.1              &\cellcolor{yellow}-    &\cellcolor{yellow}-    &\cellcolor{yellow}- &\cellcolor{yellow}-               \\
&\cellcolor{yellow}\text{CompoDistill}$^\dag$                          &\cellcolor{yellow}Qwen1.5-0.5B          &\cellcolor{yellow}1.2 M                           &\cellcolor{yellow}59.1  &\cellcolor{yellow}61.1   &\cellcolor{yellow}48.2     &\cellcolor{yellow}63.8 &\cellcolor{yellow}59.6 &\cellcolor{yellow}54.9                 &\cellcolor{yellow}85.7 &\cellcolor{yellow}\textbf{34.0}    &\cellcolor{yellow}62.9 &\cellcolor{yellow}58.3     \\
&\cellcolor{yellow}LLaVA-KD                          &\cellcolor{yellow}Qwen1.5-0.5B          &\cellcolor{yellow}1.2 M                            &\cellcolor{yellow}59.6  &\cellcolor{yellow}60.6   &\cellcolor{yellow}49.9      &\cellcolor{yellow}64.5  &\cellcolor{yellow}60.1 &\cellcolor{yellow}55.5                 &\cellcolor{yellow}85.9  &\cellcolor{yellow}30.2    &\cellcolor{yellow}63.4 &\cellcolor{yellow}58.2      \\
&\cellcolor{yellow}LLaVA-KD                          &\cellcolor{yellow}Qwen2.5-0.5B          &\cellcolor{yellow}1.2 M                            &\cellcolor{yellow}\underline{59.8}  &\cellcolor{yellow}60.6   &\cellcolor{yellow}\underline{52.0}     &\cellcolor{yellow}64.7 &\cellcolor{yellow}\underline{61.3} &\cellcolor{yellow}\underline{57.0}                 &\cellcolor{yellow}86.4 &\cellcolor{yellow}28.3    &\cellcolor{yellow}\underline{64.1} &\cellcolor{yellow}\underline{58.7}     \\
&\cellcolor{green}\textbf{\method}                          &\cellcolor{green}Qwen1.5-0.5B         &\cellcolor{green}1.2 M                            &\cellcolor{green}59.5     &\cellcolor{green}\underline{62.1}      &\cellcolor{green}48.9        &\cellcolor{green}65.2    &\cellcolor{green}60.9    &\cellcolor{green}54.6                 &\cellcolor{green}\underline{86.5}    &\cellcolor{green}30.9    &\cellcolor{green}63.8 &\cellcolor{green}58.5                \\
&\cellcolor{green}\textbf{\method}                          &\cellcolor{green}Qwen2.5-0.5B         &\cellcolor{green}1.2 M                         &\cellcolor{green}\textbf{60.2}     &\cellcolor{green}\textbf{62.7}      &\cellcolor{green}\underline{52.0}        &\cellcolor{green}\textbf{68.1}    &\cellcolor{green}\textbf{61.7}    &\cellcolor{green}\textbf{57.7}                 &\cellcolor{green}\textbf{87.6}    &\cellcolor{green}31.3    &\cellcolor{green}\textbf{65.3} &\cellcolor{green}\textbf{60.2}                  \\
\Xhline{1.5pt}
\end{tabular}
}
\label{table:main_results}
\vspace{-15pt}
\end{table*}

\subsection{Main Results}
\subsubsection{General VQA benchmarks.} \label{sec:results_vqa}
Table \ref{table:main_results} demonstrates that \method~consistently improves performance over the PT-SFT baseline across all tested backbones and student scales. The performance gains are most significant in the $\le$ 0.5B regime. For the Qwen2.5-0.5B student, \method~increases $Avg_6$ from 62.4 to 65.3 and $Avg_8$ from 57.1 to 60.2 compared to the PT-SFT baseline. This result also surpasses LLaVA-KD on the same backbone by 1.2 points in $Avg_6$ and 1.5 points in $Avg_8$. Similar robust improvements are maintained for the Qwen1.5-0.5B backbone, where \method~achieves the highest scores in both average metrics.

In the $\le$ 2B regime, \method~continues to exhibit superior performance. Using the Qwen2.5-1.5B backbone, \method~achieves 70.8 in $Avg_6$ and 66.6 in $Avg_8$, outperforming both the TinyLLaVA baseline and LLaVA-KD. Notably, \method~surpasses LLaVA-MoD (68.8 in $Avg_6$) despite using 76\% fewer training samples (1.2M vs. 5M).

Compared to CompoDistill, which relies on uniform prompt-conditioned attention supervision, \method~achieves stronger overall results by transitioning to a decomposed token-level objective targeting response-conditioned attention, e.g., improving $Avg_8$ from 61.9 to 62.9 on Qwen1.5-1.8B. This shift indicates that grounding cues manifested during decoding are particularly informative for distillation when supervised at token-level granularity. By leveraging this signal in the response phase, \method~provides a more targeted supervision pathway, yielding robust improvements in downstream VQA across backbones.

\vspace{-10pt}

\subsubsection{Compositional Reasoning (CR) benchmarks.} \label{sec:results_cr}
Table~\ref{table:compo_result} reports reasoning performance on SugarCrepe, BiVLC, and Winoground. \method~substantially strengthens compositional robustness, particularly in low-capacity regimes. For Qwen1.5-0.5B and 1.8B, \method~improves the average scores to 70.3 and 79.8, exceeding the strongest prior baseline, CompoDistill, by +1.2 and +3.0 points, respectively. Similar gains are observed in the Qwen2.5 family, where the 1.5B student reaches an average of 85.0, narrowing the gap with the 3B teacher to 0.9 points. These results highlight that token-wise guidance during decoding enhances the student’s ability to ground compositional judgments in the right visual evidence, leading to robust gains on compositional reasoning benchmarks.

\vspace{-10pt}

\subsubsection{Backbone Generalization.}
We further evaluate \method~on the MobileLLaMA backbone, where it improves the 1.4B student over the PT-SFT baseline by $+3.0$ VQA Avg and $+7.3$ CR Avg. Detailed results are provided in Appendix~\ref{appendix:additional_results}.

\begin{table*}[t]
\centering
\caption{\textbf{Compositional reasoning benchmark comparison of \method.} Performance on three compositional reasoning benchmarks across Qwen1.5 and Qwen2.5 backbones at multiple parameter scales. Models are grouped by backbone and size. Within each backbone/size group, the best result is highlighted in \textbf{bold}.
\sethlcolor{blue}\hl{Blue} indicates the teacher model for each backbone, \sethlcolor{gray}\hl{gray} rows represent models trained with the standard PT-SFT pipeline, \sethlcolor{yellow}\hl{yellow} denotes prior distillation baselines, and \sethlcolor{green}\hl{green} indicates our proposed method.}
\setlength{\tabcolsep}{3.5pt}
\renewcommand{\arraystretch}{0.78}
\footnotesize
\resizebox{0.94\linewidth}{!}{
\begin{tabular}{c|c|c|*{3}{wc{1.8cm}}|c}
\noalign{\hrule height 0.15em}
\multirow{2}{*}{LLM} & \multirow{2}{*}{Method} & \multirow{2}{*}{Size} & \multicolumn{3}{c|}{Compositional Reasoning Benchmarks} & \multirow{2}{*}{$Avg$} \\\cline{4-6}
&                    &                 & SugarCrepe  & BiVLC & Winoground &  \\\hline
\multirow{11}{*}{\rotatebox{90}{Qwen1.5}} 
&\cellcolor{blue}TinyLLaVA           &\cellcolor{blue}4B       &\cellcolor{blue}87.3        &\cellcolor{blue}93.2  &\cellcolor{blue}70.1       &\cellcolor{blue}83.5 \\ \cdashline{2-7}[1pt/1pt]
&\cellcolor{gray}TinyLLaVA           &\cellcolor{gray}                          & \cellcolor{gray}74.8        & \cellcolor{gray}83.8  & \cellcolor{gray}62.6       & \cellcolor{gray}73.7 \\
&KD                  &                          & 76.1        & 85.3  & 61.5       & 74.3 \\
&\cellcolor{yellow}LLaVA-KD            &\cellcolor{yellow}                          & \cellcolor{yellow}76.6        & \cellcolor{yellow}85.5  & \cellcolor{yellow}60.9       & \cellcolor{yellow}74.3 \\
&\cellcolor{yellow}CompoDistill        &\cellcolor{yellow}                          & \cellcolor{yellow}81.7        & \cellcolor{yellow}86.5  & \cellcolor{yellow}62.2       & \cellcolor{yellow}76.8    \\
&\cellcolor{green}\textbf{\method}    &\cellcolor{green}\multirow{-5}{*}{1.8B}     &\cellcolor{green}\textbf{83.1}           &\cellcolor{green}\textbf{89.1}     &\cellcolor{green}\textbf{67.2}          &\cellcolor{green}\textbf{79.8}    \\ \cdashline{2-7}[1pt/1pt]
&\cellcolor{gray}TinyLLaVA           &\cellcolor{gray}                          & \cellcolor{gray}52.8        & \cellcolor{gray}56.5  & \cellcolor{gray}50.2       & \cellcolor{gray}53.2 \\
&KD                  &                          & 51.0        & 50.1  & 52.1       & 51.0 \\
&\cellcolor{yellow}LLaVA-KD            &\cellcolor{yellow}                          & \cellcolor{yellow}66.9        & \cellcolor{yellow}74.5  & \cellcolor{yellow}51.7       & \cellcolor{yellow}64.4 \\
&\cellcolor{yellow}CompoDistill        &\cellcolor{yellow}                          & \cellcolor{yellow}72.1        & \cellcolor{yellow}78.6  & \cellcolor{yellow}56.6       & \cellcolor{yellow}69.1    \\
&\cellcolor{green}\textbf{\method}    &\cellcolor{green}\multirow{-5}{*}{0.5B}     &\cellcolor{green}\textbf{73.4}        &\cellcolor{green}\textbf{80.2}  &\cellcolor{green}\textbf{57.3}       &\cellcolor{green}\textbf{70.3} \\

\noalign{\hrule height 0.1em}

\multirow{5}{*}{\rotatebox{90}{Qwen2.5}} 
&\cellcolor{blue}TinyLLaVA           &\cellcolor{blue}3B       &\cellcolor{blue}88.1        &\cellcolor{blue}94.6  &\cellcolor{blue}75.0       &\cellcolor{blue}85.9 \\ \cdashline{2-7}[1pt/1pt]
&\cellcolor{gray}TinyLLaVA           &\cellcolor{gray}                          &\cellcolor{gray}\textbf{87.9}        &\cellcolor{gray}93.3  &\cellcolor{gray}70.3       &\cellcolor{gray}83.8 \\
&\cellcolor{green}\textbf{\method}    &\cellcolor{green}\multirow{-2}{*}{1.5B}     &\cellcolor{green}87.3           &\cellcolor{green}\textbf{93.8}     &\cellcolor{green}\textbf{73.9}          &\cellcolor{green}\textbf{85.0}    \\ \cdashline{2-7}[1pt/1pt]
&\cellcolor{gray}TinyLLaVA           &\cellcolor{gray}                          &\cellcolor{gray}75.1        &\cellcolor{gray}85.1  &\cellcolor{gray}57.0       &\cellcolor{gray}72.4 \\
&\cellcolor{green}\textbf{\method}    &\cellcolor{green}\multirow{-2}{*}{0.5B}     &\cellcolor{green}\textbf{78.9}           &\cellcolor{green}\textbf{87.1}     &\cellcolor{green}\textbf{59.6}          &\cellcolor{green}\textbf{75.2}    \\
\noalign{\hrule height 0.15em}
\end{tabular}
}
\label{table:compo_result}
\vspace{-5pt}
\end{table*}

\begin{wrapfigure}{r}{0.55\textwidth}
    \centering
    \vspace{-25pt}
    \captionsetup{skip=2.0pt}
    \includegraphics[width=1\linewidth]{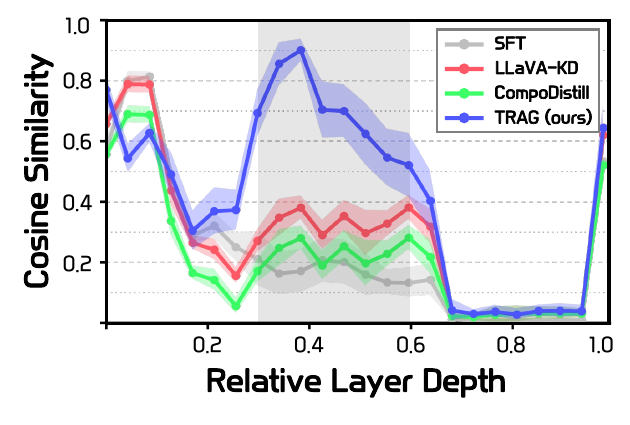}
    \caption{Layer-wise Response-to-Vision attention fidelity.}
    \label{fig:response_similarity}
    \vspace{-20pt}
\end{wrapfigure}

\subsection{Analysis of Attention Fidelity} \label{sec:exp_r2v}

We evaluate attention fidelity on \textit{VIGC-InstData}~\cite{vigc} using layer-wise teacher–student $(A_{R\rightarrow V})$ cosine similarity during autoregressive generation (without teacher forcing), following the similarity definition in Section~\ref{sec:obs_response} (Figure~\ref{fig:response_similarity}). At the 30-60\% layer depth, where cross-modal grounding is most intensive~\cite{compodistill,focus_area}, \method~achieves an attention fidelity of approximately 0.72, significantly surpassing the SFT baseline (0.18) as well as LLaVA-KD (0.32) and CompoDistill (0.41). This elevated fidelity indicates that by supervising response-conditioned attention at the token level, \method~successfully replicates the teacher's grounding logic throughout the generative process. Notably, this internal consistency directly correlates with the performance gains observed in Tables \ref{table:main_results} and \ref{table:compo_result}. These results confirm that maintaining high fidelity to the teacher's internal attention patterns is fundamental to achieving superior visual understanding and compositional robustness in low-capacity models.

\subsection{Ablation Study}
We report the main ablations on query-token selection and attention matching below. Additional ablations on relative-depth selection and response-supervision granularity are provided in Appendix~\ref{appendix:additional_results}.

\begin{wraptable}{r}{0.5\textwidth}
\vspace{-30pt}
\scriptsize
\centering
\renewcommand{\arraystretch}{0.8}
\caption{Ablation results on query token types for attention extraction.}
\setlength{\tabcolsep}{4pt}
\resizebox{1.0\linewidth}{!}{
\begin{tabular}{c|cc} 
\toprule[0.15em]
\textbf{Query Token Type} & \textbf{VQA $Avg$} & \textbf{CR $Avg$} \\
\midrule[0.1em]
Prompt             & 58.6     & 69.2        \\
Prompt + Response  & 58.4     & 71.2        \\
\rowcolor{green} \textbf{Response (Ours)} & \textbf{60.2} & \textbf{75.2} \\
\bottomrule[0.15em]
\end{tabular}
}
\label{tab:ablation_query}
\vspace{-15pt}
\end{wraptable}

\subsubsection{Query Token Type.}
Table~\ref{tab:ablation_query} compares query token choices for forming visual-attention targets. Supervising only prompt tokens yields lower averages (VQA: 58.6, CR: 69.2) than using response tokens (VQA: 60.2, CR: 75.2). Using both prompt and response tokens does not improve VQA (58.4 vs.\ 58.6) and provides only a modest gain on CR (71.2 vs.\ 69.2), remaining substantially below response-only supervision. Overall, the best performance is achieved when supervision is concentrated on response tokens, suggesting that the most informative cross-modal signal emerges during response generation.

\begin{wraptable}{r}{0.5\textwidth}
\vspace{-30pt}
\small
\centering
\renewcommand{\arraystretch}{0.8}
\caption{Ablation results on attention matching objectives.}
\setlength{\tabcolsep}{4pt}
\resizebox{1.0\linewidth}{!}{
\begin{tabular}{c|cc}
\toprule[0.15em]
\textbf{Attention Matching} & \textbf{VQA $Avg$} & \textbf{CR $Avg$} \\
\midrule[0.1em]
MSE                & 58.2     & 74.4        \\
Cosine Similarity  & 58.7     & 72.5        \\
Forward KL         & 57.6     & 73.9        \\
Reverse KL         & 58.1     & 73.4        \\
JSD                & 57.9     & 71.6        \\
\cellcolor{green}\textbf{Weighted Sum KL (Ours)} &\cellcolor{green}\textbf{60.2} &\cellcolor{green}\textbf{75.2}        \\
\bottomrule[0.15em]
\end{tabular}
}
\label{tab:ablation_matching}
\vspace{-15pt}
\end{wraptable}

\subsubsection{Attention Matching Objectives.}
Table~\ref{tab:ablation_matching} compares alternative discrepancy measures for matching teacher--student response-to-vision attention. Among the fixed objectives, cosine similarity yields the strongest VQA average (58.7), while MSE performs best on CR (74.4), indicating that no single fixed measure dominates across benchmarks. The one-sided KL variants (forward or reverse) and the symmetric JSD objective also fall short of the best fixed baselines on at least one benchmark. In contrast, our entropy-aware weighted combination of forward and reverse KL achieves the best results on both VQA (60.2) and CR (75.2), improving over the strongest fixed baselines by +1.5 and +0.8, respectively. Overall, these gains suggest that effective attention distillation benefits from token-adaptive guidance rather than a globally fixed matching rule.

\subsection{Scaling Experiments Across Model Sizes} \label{sec:exp_scaling}

\begin{wraptable}{r}{0.5\textwidth}
\vspace{-35pt}
\centering
\caption{Scaling experiments with different teacher model sizes.}
\renewcommand{\arraystretch}{0.8}
\resizebox{1.0\linewidth}{!}{
\begin{tabular}{cc|c@{\hspace{0.5cm}}c}
\toprule[0.15em]
Teacher & Student & \textbf{VQA} $Avg$ & \textbf{CR} $Avg$ \\
\midrule[0.1em]
\multicolumn{2}{c|}{Qwen1.5-0.5B (SFT)}                                           & 55.8     & 53.2                     \\
\hdashline
\rule{0pt}{2ex} Qwen1.5-1.8B                              & \multirow{2}{*}{Qwen1.5-0.5B}        & 57.3     & 63.0                     \\
Qwen1.5-4B                                &                                      & 58.5     & 70.3                     \\
\midrule[0.1em]
\multicolumn{2}{c|}{Qwen2.5-0.5B (SFT)}                                           & 57.1     & 72.4                     \\
\hdashline
\rule{0pt}{2ex} Qwen2.5-1.5B                              & \multirow{2}{*}{Qwen2.5-0.5B}        & 58.3     & 74.3                     \\
Qwen2.5-3B                                &                                      & 60.2     & 75.2                     \\
\bottomrule[0.15em]
\end{tabular}
}
\label{tab:scaling}
\vspace{-30pt}
\end{wraptable}

Table \ref{tab:scaling} examines the impact of teacher capacity on distillation performance for a fixed 0.5B student model. Across both Qwen1.5 and Qwen2.5 backbones, \method~consistently improves over the SFT baseline, with larger teachers yielding greater gains, especially on compositional reasoning. For Qwen1.5, CR Avg rises from 53.2 to 63.0 and 70.3 as the teacher scales from SFT to 1.8B and 4B; for Qwen2.5, it similarly increases from 72.4 to 74.3 and 75.2 as the teacher scales from SFT to 1.5B and 3B. These results indicate that \method~effectively captures and transfers the enhanced grounding capabilities of larger teachers, demonstrating favorable scalability with respect to teacher model size.

\section{Conclusion}
\label{conclusion}
We presented \method, a response-token-wise attention guidance objective for distilling MLLMs. By shifting cross-modal supervision from Prompt-to-Vision to Response-to-Vision attention, \method~directly targets visual grounding during generation. It further decomposes supervision into token-level objectives and adaptively balances forward and reverse KL divergences according to the teacher's attention entropy. Integrated into a standard MLLM distillation pipeline, \method~consistently improves performance on general VQA and compositional reasoning benchmarks while also improving teacher-student attention fidelity.


\section*{Acknowledgements}
This work was supported by Samsung Electronics Co., Ltd (IO251223-14934-01) and the Institute of Information \& communications Technology Planning \& Evaluation (IITP) grant funded by the Korea government(MSIT) (No.RS-2022-II220184, Development and Study of AI Technologies to Inexpensively Conform to Evolving Policy on Ethics, and No.RS-2021-II211381, Development of Causal AI through Video Understanding and Reinforcement Learning, and Its Applications to Real Environments).
%
%

\bibliographystyle{splncs04}
\bibliography{main}

@String(CVPR  = {IEEE Conf. Comput. Vis. Pattern Recog.})

@String(AAAI  = {AAAI})

@String(CVPR  = {CVPR})

@inproceedings{intermediate_layer,
  title={Towards interpreting visual information processing in vision-language models},
  author={Neo, Clement and Ong, Luke and Torr, Philip and Geva, Mor and Krueger, David and Barez, Fazl},
  booktitle={International Conference on Learning Representations},
  volume={2025},
  pages={57172--57189},
  year={2025}
}

@inproceedings{vila,
  title={Vila: On pre-training for visual language models},
  author={Lin, Ji and Yin, Hongxu and Ping, Wei and Molchanov, Pavlo and Shoeybi, Mohammad and Han, Song},
  booktitle={Proceedings of the IEEE/CVF conference on computer vision and pattern recognition},
  pages={26689--26699},
  year={2024}
}

@misc{qwen3vl,
      title={Qwen3-VL Technical Report}, 
      author={Shuai Bai and Yuxuan Cai and Ruizhe Chen and Keqin Chen and Xionghui Chen and Zesen Cheng and Lianghao Deng and Wei Ding and Chang Gao and Chunjiang Ge and Wenbin Ge and Zhifang Guo and Qidong Huang and Jie Huang and Fei Huang and Binyuan Hui and Shutong Jiang and Zhaohai Li and Mingsheng Li and Mei Li and Kaixin Li and Zicheng Lin and Junyang Lin and Xuejing Liu and Jiawei Liu and Chenglong Liu and Yang Liu and Dayiheng Liu and Shixuan Liu and Dunjie Lu and Ruilin Luo and Chenxu Lv and Rui Men and Lingchen Meng and Xuancheng Ren and Xingzhang Ren and Sibo Song and Yuchong Sun and Jun Tang and Jianhong Tu and Jianqiang Wan and Peng Wang and Pengfei Wang and Qiuyue Wang and Yuxuan Wang and Tianbao Xie and Yiheng Xu and Haiyang Xu and Jin Xu and Zhibo Yang and Mingkun Yang and Jianxin Yang and An Yang and Bowen Yu and Fei Zhang and Hang Zhang and Xi Zhang and Bo Zheng and Humen Zhong and Jingren Zhou and Fan Zhou and Jing Zhou and Yuanzhi Zhu and Ke Zhu},
      year={2025},
      eprint={2511.21631},
      archivePrefix={arXiv},
      primaryClass={cs.CV},
      journal = {arXiv preprint arXiv:2511.21631}, 
}

@article{deepseekvl,
  title={Deepseek-vl: towards real-world vision-language understanding},
  author={Lu, Haoyu and Liu, Wen and Zhang, Bo and Wang, Bingxuan and Dong, Kai and Liu, Bo and Sun, Jingxiang and Ren, Tongzheng and Li, Zhuoshu and Yang, Hao and others},
  journal={arXiv preprint arXiv:2403.05525},
  year={2024}
}

@article{llava,
  title={Visual instruction tuning},
  author={Liu, Haotian and Li, Chunyuan and Wu, Qingyang and Lee, Yong Jae},
  journal={Advances in neural information processing systems},
  volume={36},
  pages={34892--34916},
  year={2023}
}

@inproceedings{siglip,
  title={Sigmoid loss for language image pre-training},
  author={Zhai, Xiaohua and Mustafa, Basil and Kolesnikov, Alexander and Beyer, Lucas},
  booktitle={Proceedings of the IEEE/CVF international conference on computer vision},
  pages={11975--11986},
  year={2023}
}

@article{gelu,
  title={Gaussian Error Linear Units (Gelus)},
  author={Hendrycks, D},
  journal={arXiv preprint arXiv:1606.08415},
  year={2016}
}

@inproceedings{llava1.5,
  title={Improved baselines with visual instruction tuning},
  author={Liu, Haotian and Li, Chunyuan and Li, Yuheng and Lee, Yong Jae},
  booktitle={Proceedings of the IEEE/CVF conference on computer vision and pattern recognition},
  pages={26296--26306},
  year={2024}
}

@article{qvlm,
  title={Q-vlm: Post-training quantization for large vision-language models},
  author={Wang, Changyuan and Wang, Ziwei and Xu, Xiuwei and Tang, Yansong and Zhou, Jie and Lu, Jiwen},
  journal={Advances in Neural Information Processing Systems},
  volume={37},
  pages={114553--114573},
  year={2024}
}

@inproceedings{vispruner,
  title={Beyond text-visual attention: Exploiting visual cues for effective token pruning in vlms},
  author={Zhang, Qizhe and Cheng, Aosong and Lu, Ming and Zhang, Renrui and Zhuo, Zhiyong and Cao, Jiajun and Guo, Shaobo and She, Qi and Zhang, Shanghang},
  booktitle={Proceedings of the IEEE/CVF International Conference on Computer Vision},
  pages={20857--20867},
  year={2025}
}

@misc{hintonkd,
      title={Distilling the Knowledge in a Neural Network}, 
      author={Geoffrey Hinton and Oriol Vinyals and Jeff Dean},
      year={2015},
      eprint={1503.02531},
      archivePrefix={arXiv},
      primaryClass={stat.ML},
}

@inproceedings{
minillm,
title={Mini{LLM}: Knowledge Distillation of Large Language Models},
author={Yuxian Gu and Li Dong and Furu Wei and Minlie Huang},
booktitle={The Twelfth International Conference on Learning Representations},
year={2024},
}

@inproceedings{
gkd,
title={On-Policy Distillation of Language Models: Learning from Self-Generated Mistakes},
author={Rishabh Agarwal and Nino Vieillard and Yongchao Zhou and Piotr Stanczyk and Sabela Ramos Garea and Matthieu Geist and Olivier Bachem},
booktitle={The Twelfth International Conference on Learning Representations},
year={2024},
}

@inproceedings{
distillm,
title={Disti{LLM}: Towards Streamlined Distillation for Large Language Models},
author={Jongwoo Ko and Sungnyun Kim and Tianyi Chen and Se-Young Yun},
booktitle={Forty-first International Conference on Machine Learning},
year={2024},
}

@article{llavadi,
  title={Llavadi: What matters for multimodal large language models distillation},
  author={Xu, Shilin and Li, Xiangtai and Yuan, Haobo and Qi, Lu and Tong, Yunhai and Yang, Ming-Hsuan},
  journal={arXiv preprint arXiv:2407.19409},
  year={2024}
}

@inproceedings{llavakd,
  title={Llava-kd: A framework of distilling multimodal large language models},
  author={Cai, Yuxuan and Zhang, Jiangning and He, Haoyang and He, Xinwei and Tong, Ao and Gan, Zhenye and Wang, Chengjie and Xue, Zhucun and Liu, Yong and Bai, Xiang},
  booktitle={Proceedings of the IEEE/CVF International Conference on Computer Vision},
  pages={239--249},
  year={2025}
}

@inproceedings{compodistill,
title={CompoDistill: Attention Distillation for Compositional Reasoning in Multimodal {LLM}s},
author={Jiwan Kim and Kibum Kim and Sangwoo Seo and Chanyoung Park},
booktitle={The Fourteenth International Conference on Learning Representations},
year={2026},
}

@inproceedings{llavamod,
  title={LLaVA-MoD: Making LLaVA Tiny via MoE-Knowledge Distillation},
  author={Shu, Fangxun and Liao, Yue and Zhang, Lei and Zhuo, Le and Xu, Chenning and Zhang, Guanghao and Shi, Haonan and Chan, Long and Yu, Zhelun and He, Wanggui and others},
  booktitle={The Thirteenth International Conference on Learning Representations},
  year={2025}
}

@inproceedings{alignkd,
  title={Align-KD: Distilling Cross-Modal Alignment Knowledge for Mobile Vision-Language Large Model Enhancement},
  author={Feng, Qianhan and Li, Wenshuo and Lin, Tong and Chen, Xinghao},
  booktitle={Proceedings of the Computer Vision and Pattern Recognition Conference},
  pages={4178--4188},
  year={2025}
}

@inproceedings{at,
  title={Paying More Attention to Attention: Improving the Performance of Convolutional Neural Networks via Attention Transfer},
  author={Zagoruyko, Sergey and Komodakis, Nikos},
  booktitle={International Conference on Learning Representations},
  year={2017}
}

@article{aftkd,
  title={Attention and feature transfer based knowledge distillation},
  author={Yang, Guoliang and Yu, Shuaiying and Sheng, Yangyang and Yang, Hao},
  journal={Scientific Reports},
  volume={13},
  number={1},
  pages={18369},
  year={2023},
  publisher={Nature Publishing Group UK London}
}

@article{sakd,
  title={SAKD: Sparse attention knowledge distillation},
  author={Guo, Zhen and Zhang, Pengzhou and Liang, Peng},
  journal={Image and Vision Computing},
  volume={146},
  pages={105020},
  year={2024},
  publisher={Elsevier}
}

@inproceedings{cacr,
  title={Cross-modal attention congruence regularization for vision-language relation alignment},
  author={Pandey, Rohan and Shao, Rulin and Liang, Paul Pu and Salakhutdinov, Ruslan and Morency, Louis-Philippe},
  booktitle={Proceedings of the 61st Annual Meeting of the Association for Computational Linguistics (Volume 1: Long Papers)},
  pages={5444--5455},
  year={2023}
}

@inproceedings{catkd,
  title={Class attention transfer based knowledge distillation},
  author={Guo, Ziyao and Yan, Haonan and Li, Hui and Lin, Xiaodong},
  booktitle={Proceedings of the IEEE/CVF conference on computer vision and pattern recognition},
  pages={11868--11877},
  year={2023}
}

@inproceedings{gqa,
  title={Gqa: A new dataset for real-world visual reasoning and compositional question answering},
  author={Hudson, Drew A and Manning, Christopher D},
  booktitle={Proceedings of the IEEE/CVF conference on computer vision and pattern recognition},
  pages={6700--6709},
  year={2019}
}

@article{sqa,
  title={Learn to explain: Multimodal reasoning via thought chains for science question answering},
  author={Lu, Pan and Mishra, Swaroop and Xia, Tanglin and Qiu, Liang and Chang, Kai-Wei and Zhu, Song-Chun and Tafjord, Oyvind and Clark, Peter and Kalyan, Ashwin},
  journal={Advances in Neural Information Processing Systems},
  volume={35},
  pages={2507--2521},
  year={2022}
}

@inproceedings{textvqa,
  title={Towards vqa models that can read},
  author={Singh, Amanpreet and Natarajan, Vivek and Shah, Meet and Jiang, Yu and Chen, Xinlei and Batra, Dhruv and Parikh, Devi and Rohrbach, Marcus},
  booktitle={Proceedings of the IEEE/CVF conference on computer vision and pattern recognition},
  pages={8317--8326},
  year={2019}
}

@inproceedings{mme,
  title={Mme: A comprehensive evaluation benchmark for multimodal large language models},
  author={Fu, Chaoyou and Chen, Peixian and Shen, Yunhang and Qin, Yulei and Zhang, Mengdan and Lin, Xu and Yang, Jinrui and Zheng, Xiawu and Li, Ke and Sun, Xing and others},
  booktitle={The Thirty-ninth Annual Conference on Neural Information Processing Systems Datasets and Benchmarks Track},
  year={2025}
}

@inproceedings{mmb,
  title={Mmbench: Is your multi-modal model an all-around player?},
  author={Liu, Yuan and Duan, Haodong and Zhang, Yuanhan and Li, Bo and Zhang, Songyang and Zhao, Wangbo and Yuan, Yike and Wang, Jiaqi and He, Conghui and Liu, Ziwei and others},
  booktitle={European conference on computer vision},
  pages={216--233},
  year={2024},
  organization={Springer}
}

@inproceedings{pope,
  title={Evaluating Object Hallucination in Large Vision-Language Models},
  author={Li, Yifan and Du, Yifan and Zhou, Kun and Wang, Jinpeng and Zhao, Xin and Wen, Ji-Rong},
  booktitle={The 2023 Conference on Empirical Methods in Natural Language Processing}
}

@inproceedings{mmmu,
  title={Mmmu: A massive multi-discipline multimodal understanding and reasoning benchmark for expert agi},
  author={Yue, Xiang and Ni, Yuansheng and Zhang, Kai and Zheng, Tianyu and Liu, Ruoqi and Zhang, Ge and Stevens, Samuel and Jiang, Dongfu and Ren, Weiming and Sun, Yuxuan and others},
  booktitle={Proceedings of the IEEE/CVF Conference on Computer Vision and Pattern Recognition},
  pages={9556--9567},
  year={2024}
}

@article{sugarcrepe,
  title={Sugarcrepe: Fixing hackable benchmarks for vision-language compositionality},
  author={Hsieh, Cheng-Yu and Zhang, Jieyu and Ma, Zixian and Kembhavi, Aniruddha and Krishna, Ranjay},
  journal={Advances in neural information processing systems},
  volume={36},
  pages={31096--31116},
  year={2023}
}

@article{bivlc,
  title={Bivlc: Extending vision-language compositionality evaluation with text-to-image retrieval},
  author={Miranda, Imanol and Salaberria, Ander and Agirre, Eneko and Azkune, Gorka},
  journal={Advances in Neural Information Processing Systems},
  volume={37},
  pages={101880--101904},
  year={2024}
}

@inproceedings{winoground,
  title={Winoground: Probing vision and language models for visio-linguistic compositionality},
  author={Thrush, Tristan and Jiang, Ryan and Bartolo, Max and Singh, Amanpreet and Williams, Adina and Kiela, Douwe and Ross, Candace},
  booktitle={Proceedings of the IEEE/CVF Conference on Computer Vision and Pattern Recognition},
  pages={5238--5248},
  year={2022}
}

@inproceedings{vigc,
  title={Vigc: Visual instruction generation and correction},
  author={Wang, Bin and Wu, Fan and Han, Xiao and Peng, Jiahui and Zhong, Huaping and Zhang, Pan and Dong, Xiaoyi and Li, Weijia and Li, Wei and Wang, Jiaqi and others},
  booktitle={Proceedings of the AAAI Conference on Artificial Intelligence},
  volume={38},
  number={6},
  pages={5309--5317},
  year={2024}
}

@article{focus_area,
  title={Why is spatial reasoning hard for vlms? an attention mechanism perspective on focus areas},
  author={Chen, Shiqi and Zhu, Tongyao and Zhou, Ruochen and Zhang, Jinghan and Gao, Siyang and Niebles, Juan Carlos and Geva, Mor and He, Junxian and Wu, Jiajun and Li, Manling},
  journal={arXiv preprint arXiv:2503.01773},
  year={2025}
}

@article{transformers,
  title={Attention is all you need},
  author={Vaswani, Ashish and Shazeer, Noam and Parmar, Niki and Uszkoreit, Jakob and Jones, Llion and Gomez, Aidan N and Kaiser, {\L}ukasz and Polosukhin, Illia},
  journal={Advances in neural information processing systems},
  volume={30},
  year={2017}
}

@inproceedings{akl,
  title={Rethinking kullback-leibler divergence in knowledge distillation for large language models},
  author={Wu, Taiqiang and Tao, Chaofan and Wang, Jiahao and Yang, Runming and Zhao, Zhe and Wong, Ngai},
  booktitle={Proceedings of the 31st International Conference on Computational Linguistics},
  pages={5737--5755},
  year={2025}
}

@article{bdkd,
  title={Bd-kd: Balancing the divergences for online knowledge distillation},
  author={Amara, Ibtihel and Sepahvand, Nazanin and Meyer, Brett H and Gross, Warren J and Clark, James J},
  journal={arXiv preprint arXiv:2212.12965},
  year={2022}
}

@article{alignti,
      title={Beyond Next-Token Alignment: Distilling Multimodal Large Language Models via Token Interactions}, 
      author={Lin Chen and Xiaoke Zhao and Kun Ding and Weiwei Feng and Changtao Miao and Zili Wang and Wenxuan Guo and Ying Wang and Kaiyuan Zheng and Bo Zhang and Zhe Li and Shiming Xiang},
      journal={arXiv preprint arXiv:2602.09483},
      year={2026},
}

@InProceedings{movekd,
    author    = {Cao, Jiajun and Zhang, Yuan and Huang, Tao and Lu, Ming and Zhang, Qizhe and An, Ruichuan and Ma, Ningning and Zhang, Shanghang},
    title     = {MoVE-KD: Knowledge Distillation for VLMs with Mixture of Visual Encoders},
    booktitle = {Proceedings of the Computer Vision and Pattern Recognition Conference (CVPR)},
    month     = {June},
    year      = {2025},
    pages     = {19846-19856}
}

@article{inter1,
  title={Attention re-alignment in multimodal large language models via intermediate-layer guidance},
  author={Chen, Yanming and Wang, Pandong and Qin, Guofeng and Wu, Wei and Chen, Ming and Hao, Yongtao},
  journal={Scientific Reports},
  year={2026},
  publisher={Nature Publishing Group UK London}
}

@inproceedings{inter2,
  title={What do visual tokens really encode? uncovering sparsity and redundancy in multimodal large language models},
  author={Fan, Yingqi and Tong, Junlong and Zhao, Anhao and Shen, Xiaoyu},
  booktitle={Proceedings of the IEEE/CVF Conference on Computer Vision and Pattern Recognition},
  pages={11987--11997},
  year={2026}
}

@inproceedings{
vgs,
title={Decomposed On-Policy Distillation for Vision-Language Reasoning: Steering Gradients for Visual Grounding},
author={Hee Suk Yoon and Eunseop Yoon and Jaehyun Jang and SooHwan Eom and Ji Woo Hong and Mark A. Hasegawa-Johnson and Qi Dai and Chong Luo and Chang D. Yoo},
booktitle={Forty-third International Conference on Machine Learning},
year={2026},
}

@inproceedings{pdcr,
  title={PDCR: Perception-Decomposed Confidence Reward for Vision-Language Reasoning},
  author={Yoon, Hee Suk and Yoon, Eunseop and Hong, Ji Woo and Eom, SooHwan and Koo, Gwanhyeong and Hasegawa-Johnson, Mark and Dai, Qi and Luo, Chong and Yoo, Chang D},
  booktitle={Proceedings of the IEEE/CVF Conference on Computer Vision and Pattern Recognition},
  pages={18881--18891},
  year={2026}
}
\clearpage

\makeatletter

\renewcommand*\l@section[2]{%
  \vspace{0.55cm}%
  \@dottedtocline{1}{0em}{3.2em}{\large\bfseries #1}{\large\bfseries #2}%
}

\renewcommand*\l@subsection[2]{%
  \vspace{0.18cm}%
  \@dottedtocline{2}{2.8em}{4.0em}{\normalsize #1}{\normalsize #2}%
}

\newcommand{\supplementarytoc}{%
  \begingroup
  \hypersetup{colorlinks=true, linkcolor=black}%
  \setcounter{tocdepth}{2}%
  \begin{center}
    \begin{minipage}{0.92\textwidth}
      \normalsize
      \@starttoc{atoc}%
    \end{minipage}
  \end{center}
  \endgroup
}

\let\origsection\section
\let\origsubsection\subsection

\newif\ifinappendixtoc
\inappendixtocfalse

\RenewDocumentCommand{\section}{s o m}{%
  \IfBooleanTF{#1}{%
    \IfNoValueTF{#2}
      {\origsection*{#3}}
      {\origsection*[#2]{#3}}%
  }{%
    \IfNoValueTF{#2}
      {\origsection{#3}}
      {\origsection[#2]{#3}}%
    \ifinappendixtoc
      \addcontentsline{atoc}{section}{%
        \protect\numberline{\thesection}\IfNoValueTF{#2}{#3}{#2}%
      }%
    \fi
  }%
}

\RenewDocumentCommand{\subsection}{s o m}{%
  \IfBooleanTF{#1}{%
    \IfNoValueTF{#2}
      {\origsubsection*{#3}}
      {\origsubsection*[#2]{#3}}%
  }{%
    \IfNoValueTF{#2}
      {\origsubsection{#3}}
      {\origsubsection[#2]{#3}}%
    \ifinappendixtoc
      \addcontentsline{atoc}{subsection}{%
        \protect\numberline{\thesubsection}\IfNoValueTF{#2}{#3}{#2}%
      }%
    \fi
  }%
}

\makeatother

\newcommand{\makesupplementarytitle}{%
  \begin{center}
    \vspace*{0.1cm}

    {\LARGE\bfseries \textit{Supplementary Material for}\par}

    \vspace{0.3cm}

    {\Large\bfseries
    Token-level Response-visual Attention Guidance\\[0.15cm]
    for Multimodal LLMs Knowledge Distillation\par}

    \vspace{0.3cm}

    \rule{1.0\textwidth}{0.5pt}

    \vspace{0.35cm}

    {\large\bfseries Contents\par}

    \vspace{0.2cm}
  \end{center}
}

\appendix

\renewcommand{\theHsection}{appendix.\Alph{section}}
\renewcommand{\theHsubsection}{appendix.\Alph{section}.\arabic{subsection}}
\renewcommand{\theHsubsubsection}{appendix.\Alph{section}.\arabic{subsection}.\arabic{subsubsection}}

\inappendixtoctrue
\makesupplementarytitle
\supplementarytoc
\clearpage

\section{Implementation Details}
\label{appendix:implementation}
This section provides the comprehensive training configurations and hyperparameters for the three-stage training pipeline of \method.

As detailed in Table \ref{tab:appendix_additional_imple}, we utilize the SigLIP-SO400M/14 vision encoder with an input image resolution of 384$\times$384, yielding a vision sequence length of 729. The maximum sequence length for the LLM is set to 2048. The vision encoder remains frozen across all three training stages.

During the Distilled Pre-Training (DPT) stage, only the projection module is trained. We use the \textit{LLaVA-Pretrain-558K} dataset for this stage, utilizing a learning rate of 1e-3 and a global batch size of 256. In the subsequent Supervised Fine-Tuning (SFT) and Distilled Fine-Tuning (DFT) stages, both the projection module and the LLM backbone are fully finetuned. Both stages use the \textit{LLaVA-Instruct-665K} dataset and employ a reduced learning rate of 2e-5 with a global batch size of 128.

Across all stages, training is conducted for 1.0 epoch using the AdamW optimizer ($\beta_1 = 0.9, \beta_2 = 0.99$). The learning rate follows a cosine decay schedule with a warm-up ratio of 0.03, and weight decay is set to 0.0. All models are trained using BFloat16 numerical precision and DeepSpeed ZeRO-2 optimization.

\begin{table*}[h]
\centering
\caption{Training hyperparameters for each stage in our three-stage training pipeline. A checkmark \cmark~indicates that the corresponding module is updated in that stage, while a cross \xmark~indicates that it is frozen. In distilled pre-training, only the projection module is optimized. In supervised fine-tuning and distilled fine-tuning, the vision encoder is kept frozen while the projection module and the LLM backbone are optimized.}
\label{tab:appendix_additional_imple}

\setlength{\aboverulesep}{-1pt}
\setlength{\belowrulesep}{0pt}
\renewcommand{\arraystretch}{1.2}
\setlength{\tabcolsep}{7pt}
\resizebox{\linewidth}{!}{
\begin{tabular}{l|ccc}
\toprule[0.15em]
\multicolumn{1}{l}{\multirow{2}{*}{\textbf{Configuration}}}
& \textbf{Distilled}
& \textbf{Supervised}
& \textbf{Distilled} \\
\multicolumn{1}{c}{}
& \textbf{Pre-Training}
& \textbf{Fine-Tuning}
& \textbf{Fine-Tuning} \\
\midrule[0.1em]
Vision Encoder     & \xmark & \xmark & \xmark \\
Projection Module  & \cmark & \cmark & \cmark \\
LLM Backbone       & \xmark & \cmark & \cmark \\
\midrule
Training data                   & \textit{LLaVA-Pretrain-558K} & \textit{LLaVA-Instruct-665K} & \textit{LLaVA-Instruct-665K} \\
Vision Encoder                  & \multicolumn{3}{c}{SigLIP-SO400M/14@384} \\
Image resolution                & \multicolumn{3}{c}{384 $\times$ 384} \\
Vision encoder sequence length  & \multicolumn{3}{c}{729} \\
LLM sequence length             & \multicolumn{3}{c}{2048} \\
Optimizer                       & \multicolumn{3}{c}{AdamW} \\
Optimizer hyperparameters      & \multicolumn{3}{c}{$\beta_1 = 0.9,\ \beta_2 = 0.99$} \\
Learning rate                   & $1\text{e}{-3}$ & $2\text{e}{-5}$ & $2\text{e}{-5}$ \\
Learning rate scheduler         & \multicolumn{3}{c}{Cosine decay} \\
Weight decay                    & \multicolumn{3}{c}{0.0} \\
Training epochs                 & \multicolumn{3}{c}{1.0} \\
Warm-up ratio                   & \multicolumn{3}{c}{0.03} \\
Global batch size               & 256 & 128 & 128 \\
Numerical precision             & \multicolumn{3}{c}{BFloat16} \\
Model parallelism               & \multicolumn{3}{c}{ZeRO-2} \\
\bottomrule[0.15em]
\end{tabular}}
\end{table*}

\section{Evaluation Details}
\label{appendix:evaluation}
\subsection{Evaluation Setup}
\label{appendix:eval_setup}
For the general VQA benchmarks, we strictly adhere to the evaluation configurations established by LLaVA-1.5~\cite{llava1.5}. In contrast, compositional reasoning benchmarks are not directly amenable to standard VQA-style evaluation, as they are formulated as image-caption pairs rather than explicit question-answer pairs. We therefore adopt the multiple-choice evaluation protocol originally designed to assess GPT-4V in SugarCrepe~\cite{sugarcrepe}, and apply the same procedure to BiVLC~\cite{bivlc} and Winoground~\cite{winoground}.

\begin{figure}
    \centering
    \includegraphics[width=1\linewidth]{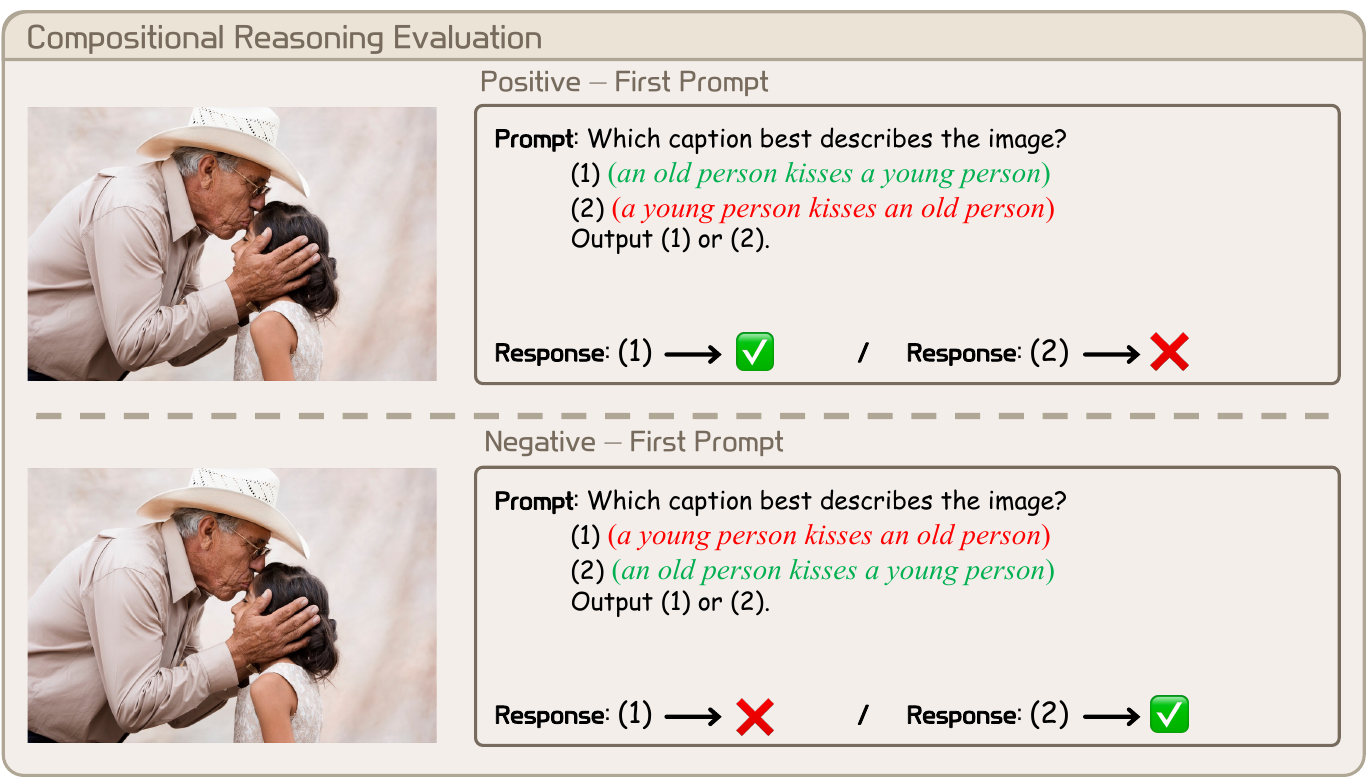}
    \caption{\textbf{Illustration of the Compositional Reasoning Evaluation setup.} The model is queried to select the correct caption index. To control for positional bias, the evaluation is performed under two swapped configurations: `\textit{Positive-First Prompt}' and `\textit{Negative-First Prompt}'.}
    \label{fig:cr_eval}
\end{figure}

Specifically, each instance in these compositional reasoning benchmarks comprises one or two images paired with a positive (matching) caption and a negative (mismatching) caption. As illustrated in Figure~\ref{fig:cr_eval}, the model is prompted to select the index of the caption that best describes the visual content. As part of this adopted protocol, to mitigate positional bias, we evaluate each instance twice by swapping the order of the candidate captions (e.g., placing the positive caption as option (1) in the `\textit{Positive-First Prompt}' setting, and as option (2) in the `\textit{Negative-First Prompt}' setting). The model's response is considered correct only if it successfully outputs the exact index corresponding to the positive caption.

\subsection{Evaluation Metrics}
\label{appendix:eval_metrics}
\subsubsection{General VQA Benchmarks.}
We report benchmark-specific metrics following the standard evaluation protocol for each respective dataset. For GQA~\cite{gqa}, TextVQA~\cite{textvqa}, and MMMU~\cite{mmmu}, accuracy serves as the primary evaluation metric. For ScienceQA~\cite{sqa}, we report accuracy exclusively on the \textit{image-context} subset to focus on multimodal evaluation. For MME~\cite{mme}, we report the official benchmark score. For MMBench and $\text{MMBench}^{\text{CN}}$~\cite{mmb}, we report the accuracy scores obtained directly from the official submission server. For POPE~\cite{pope}, we report the F1-score, aligning with prior works such as LLaVA-MoD~\cite{llavamod}.

\subsubsection{Compositional Reasoning Benchmarks.}
For SugarCrepe~\cite{sugarcrepe}, BiVLC~\cite{bivlc}, and Winoground~\cite{winoground}, we evaluate the accuracy under both the positive-first and negative-first prompt configurations. We report the average of these two accuracies for each predefined split within the benchmarks.

\section{Decomposition of the CompoDistill Objective}
\label{appendix:compo_formulation}
In this section, we revisit the attention distillation objective employed in CompoDistill~\cite{compodistill} and reinterpret its formulation by decomposing it into uni-modal and cross-modal components.

Let $\mathbf{A}_l \in \mathbb{R}^{(N+L)\times(N+L)}$ denote the self-attention matrix at decoder layer $l$, where the initial $N$ indices correspond to visual tokens and the subsequent $L$ indices correspond to textual tokens. In CompoDistill, supervision is applied to a specific attention submatrix where the queries consist of the sequence preceding the response generation (i.e., visual and prompt tokens) and the keys are restricted to the visual tokens. Formally, for each supervised layer $l$, the target attention block is defined as:

\begin{equation}
\mathbf{A}_l[1\!:\!N+L_P,\;1\!:\!N].
\label{eq:compodistill_block}
\end{equation}

Following their grouped layer matching scheme, let $\mathcal{M}_{\mathcal{S}}$ denote the set of supervised student layers, and $G_l$ denote the assigned group of teacher layers for a given student layer $l \in \mathcal{M}_{\mathcal{S}}$. The aggregated teacher attention block is computed as:

\begin{equation}
\mathbf{A}^{\mathcal{T}}_{G_l}[1\!:\!N+L_P,\;1\!:\!N]
=
\frac{1}{|G_l|}
\sum_{g\in G_l}
\mathbf{A}^{\mathcal{T}}_{g}[1\!:\!N+L_P,\;1\!:\!N].
\end{equation}

Using this notation, the original attention distillation objective in CompoDistill can be formulated as:
\begin{equation}
\mathcal{L}^{(\text{CompoDistill})}_{\text{attn}}
=
\sum_{l\in\mathcal{M}_{\mathcal S}}
\mathcal{D}_{\cos}
\!\left(
\mathbf{A}^{\mathcal{T}}_{G_l}[1\!:\!N+L_P,\;1\!:\!N],
\mathbf{A}^{\mathcal{S}}_{l}[1\!:\!N+L_P,\;1\!:\!N]
\right),
\label{eq:compodistill_attn_appendix}
\end{equation}
where \(\mathcal{D}_{\cos}(\cdot,\cdot)\) denotes the cosine distance.

This supervised attention block can be partitioned based on the modality of the queries. Specifically, the query index range $1:N+L_P$ encompasses both the visual token span ($1:N$) and the prompt token span ($N+1:N+L_P$). Consequently, the attention block can be decomposed into two distinct components:

\begin{align}
\mathbf{A}_l[1\!:\!N+L_P,\;1\!:\!N]
=
\begin{bmatrix}
\mathbf{A}_l[1\!:\!N,\;1\!:\!N] \\
\mathbf{A}_l[N+1\!:\!N+L_P,\;1\!:\!N]
\end{bmatrix}.
\end{align}

The upper submatrix, $ \mathbf{A}_l[1:N, 1:N] $, represents attention from visual queries to visual keys. As both queries and keys operate within the visual modality, this component functions strictly as a uni-modal (vision-to-vision) alignment term.

The lower submatrix, $ \mathbf{A}_l[N+1:N+L_P, 1:N] $, captures attention from prompt queries to visual keys. Following the notation established in the main paper, this corresponds precisely to the Prompt-to-Vision attention submatrix:

\begin{equation}
\mathbf{A}_l(\mathcal{I}_P)
:=
\mathbf{A}_l[\mathcal{I}_P,\;1\!:\!N],
\qquad
\mathcal{I}_P=\{N+1,\dots,N+L_P\}.
\end{equation}

Because this submatrix governs the interaction between textual queries and visual keys, it serves as the cross-modal attention alignment term.

Therefore, while CompoDistill aggregates attention supervision into a single block, this formulation inadvertently conflates two distinct functional roles: uni-modal visual self-attention and cross-modal Prompt-to-Vision attention. In this work, our primary focus lies in the cross-modal component, as it explicitly dictates how textual tokens attend to and extract visual evidence. Building upon this decomposition, \method~explicitly isolates the cross-modal interaction and shifts the supervisory signal from Prompt-to-Vision to Response-to-Vision attention, effectively providing a more targeted distillation objective.

\section{Discussion on Computational Cost}
\label{appendix:cost}

We first report empirical computational costs in Table~\ref{tab:empirical_cost}. All measurements are conducted on 2 NVIDIA A100 80GB GPUs under the same training configuration. Compared with the non-attention-based LLaVA-KD baseline, \method~incurs additional overhead due to attention extraction and token-wise adaptive KL computation. Specifically, \method~requires 52.8 hours in total and 67.3 GB peak memory per GPU, corresponding to an overhead of +3.0 hours and +17.2 GB/GPU over LLaVA-KD. However, compared with the attention-based CompoDistill baseline, \method~reduces both training time and memory usage by aggregating attention across depth before applying the adaptive KL objective, saving 2.4 hours and 3.9 GB/GPU.

\begin{table}[h]
\centering
\caption{Empirical computational cost analysis. We report peak GPU memory and stage-wise wall-clock time measured on 2 NVIDIA A100 80GB GPUs.}
\label{tab:empirical_cost}
\renewcommand{\arraystretch}{0.95}
\setlength{\tabcolsep}{5pt}
\resizebox{1.\linewidth}{!}{
\begin{tabular}{l!{\vrule width 0.8pt}c c c c !{\vrule width 0.8pt} c}
\Xhline{1.5pt}
\textbf{Method} 
& \textbf{Peak Memory (GB/GPU)} $\downarrow$ 
& \textbf{PT/DPT (h)} $\downarrow$ 
& \textbf{SFT (h)} $\downarrow$ 
& \textbf{DFT (h)} $\downarrow$ 
& \textbf{Total (h)} $\downarrow$ \\
\Xhline{1pt}
PT-SFT                    & 36.7 & 4.7  & 6.9 & -    & 11.6 \\
LLaVA-KD~\cite{llavakd}   & 50.1 & 14.9 & 6.8 & 28.1 & 49.8 \\
CompoDistill~\cite{compodistill} & 71.2 & 12.5 & 7.3 & 35.4 & 55.2 \\
\rowcolor{green}
\method~(Ours)   & 67.3 & 15.5 & 6.8 & 30.5 & 52.8 \\
\Xhline{1.5pt}
\end{tabular}
}
\end{table}

We further analyze the asymptotic computational overhead of \method~by decomposing the distillation objectives. Let $N$ denote the number of visual tokens, $L_P$ the prompt length, $L_R$ the response length, $M_S = |\mathcal{M}_{\mathcal{S}}|$ the number of intermediate student layers selected for supervision, $M_T = |\mathcal{M}_{\mathcal{T}}|$ the number of teacher layers used for aggregation in \method, and $D_\mathbf{v}$ the dimensionality of the vision logits. For CompoDistill, let $G_l$ denote the teacher layer group paired with student layer $l \in \mathcal{M}_{\mathcal{S}}$. All referenced methods share a foundational base cost comprising the standard language modeling loss and response-level logit distillation.

First, we evaluate the uni-modal distillation overhead. CompoDistill~\cite{compodistill} relies on intra-modal visual attention distillation ($\textbf{A}_{V \rightarrow V}$). This requires forming a teacher-side reference attention map for each supervised student layer by aggregating the corresponding teacher group $G_l$, followed by cosine distance computation between two $N \times N$ attention matrices. Accordingly, the resulting uni-modal complexity is
\begin{equation}
    \mathcal{O}\!\left(\left(M_S + \sum_{l \in \mathcal{M}_{\mathcal{S}}} |G_l|\right) \cdot N^2\right),
\end{equation}
where the term $M_S$ corresponds to the repeated cosine evaluations on the student side, and the term $\sum_{l \in \mathcal{M}_{\mathcal{S}}} |G_l|$ accounts for teacher-group reference formation.

In contrast, \method~adopts the uni-modal supervision strategy of LLaVA-KD~\cite{llavakd}. Crucially, the logit matching $\mathcal{O}(N \cdot D_\mathbf{v})$ and visual affinity matching $\mathcal{O}(N^2)$ are computed only once at the final output representation layer. Thus, the total uni-modal complexity for \method~is $\mathcal{O}(N^2 + N \cdot D_\mathbf{v})$, effectively eliminating the depth-wise multiplier from the uni-modal distance computation.

Second, we analyze the cross-modal distillation overhead. CompoDistill applies attention supervision from prompt queries to visual keys ($\textbf{A}_{P \rightarrow V}$). For each supervised student layer, it first forms a teacher-side prompt-to-vision reference by aggregating the corresponding teacher group $G_l$ and then computes a cosine distance loss over an $L_P \times N$ submatrix. Accordingly, the cross-modal complexity becomes
\begin{equation}
    \mathcal{O}\!\left(\left(M_S + \sum_{l \in \mathcal{M}_{\mathcal{S}}} |G_l|\right)\cdot L_P \cdot N\right).
\end{equation}

In contrast, \method~structurally decouples layer aggregation from the distance computation for its Response-to-Vision ($\textbf{A}_{R \rightarrow V}$) objective. First, \method~aggregates the student and teacher attention distributions across their respective target layer sets via simple element-wise additions, incurring
\begin{equation}
    \mathcal{O}\!\left((M_S + M_T)\cdot L_R \cdot N\right).
\end{equation}

Then, an Adaptive Kullback-Leibler (KL) objective is applied at the token level, where the teacher attention entropy determines the weighting between the forward and reverse KL terms for each response token. Crucially, this per-token evaluation is performed exactly once on the single depth-aggregated representation. Therefore, despite the granular token-level formulation, the actual adaptive distance computation complexity of \method~is $\mathcal{O}(L_R \cdot N)$ after aggregation.

In summary, CompoDistill scales both uni-modal and cross-modal distance computations with the supervised depth through repeated student-side matching and teacher-group reference formation, resulting in $\mathcal{O}((M_S + \sum_{l \in \mathcal{M}_{\mathcal{S}}} |G_l|)\cdot N^2)$ for uni-modal distillation and $\mathcal{O}((M_S + \sum_{l \in \mathcal{M}_{\mathcal{S}}} |G_l|)\cdot L_P \cdot N)$ for cross-modal distillation. In contrast, \method~restricts the heavy uni-modal distance computation to a single output-level evaluation and confines the depth-dependent part of the cross-modal branch to simple aggregation operations. The remaining Adaptive KL evaluation is performed only once on the aggregated Response-to-Vision representation. Table \ref{tab:comp_cost} clearly illustrates how \method~removes the depth multiplier from the heavy divergence evaluation itself and limits depth-dependent computation to aggregation.

\begin{table}[h]
\centering
\setlength{\aboverulesep}{-1pt}
\setlength{\belowrulesep}{0pt}
\caption{Comprehensive comparison of distillation overhead. \method~fundamentally eliminates the depth multiplier from complex distance calculations, restricting the layer-wise overhead strictly to simple aggregation.}
\label{tab:comp_cost}
\resizebox{\linewidth}{!}{
\begin{tabular}{l|cc|cc}
\toprule
\multirow{2}{*}{\textbf{Method}} & \multicolumn{2}{c|}{\textbf{Uni-modal Distillation}} & \multicolumn{2}{c}{\textbf{Cross-modal Distillation}} \\
\cmidrule{2-5}
 & \textbf{Complexity} & \textbf{Depth} & \textbf{Complexity} & \textbf{Objective Granularity} \\
\midrule
LLaVA-KD & $\mathcal{O}(N^2 + N \cdot D_\mathbf{v})$ & $1$ (Output) & - & - \\
\hdashline
CompoDistill & $\mathcal{O}((M_S + \sum_l |G_l|)\cdot N^2)$ & Layer-wise & $\mathcal{O}((M_S + \sum_l |G_l|)\cdot L_P \cdot N)$ & Block-level Cosine \\
\hdashline
\multirow{2}{*}{\textbf{\method~(Ours)}} 
& \multirow{2}{*}{$\mathcal{O}(N^2 + N \cdot D_\mathbf{v})$} 
& \multirow{2}{*}{$1$ (Output)} 
& $\mathcal{O}((M_S + M_T)\cdot L_R \cdot N)$ 
& Aggregation \\
& & 
& $+\mathcal{O}(L_R \cdot N)$ 
& + Per-token Adaptive KL \\
\bottomrule
\end{tabular}}
\end{table}

\section{Additional examples for Attention Granularity Analysis}
\label{appendix:obs2}
\begin{figure}
    \centering
    \includegraphics[width=0.8\linewidth]{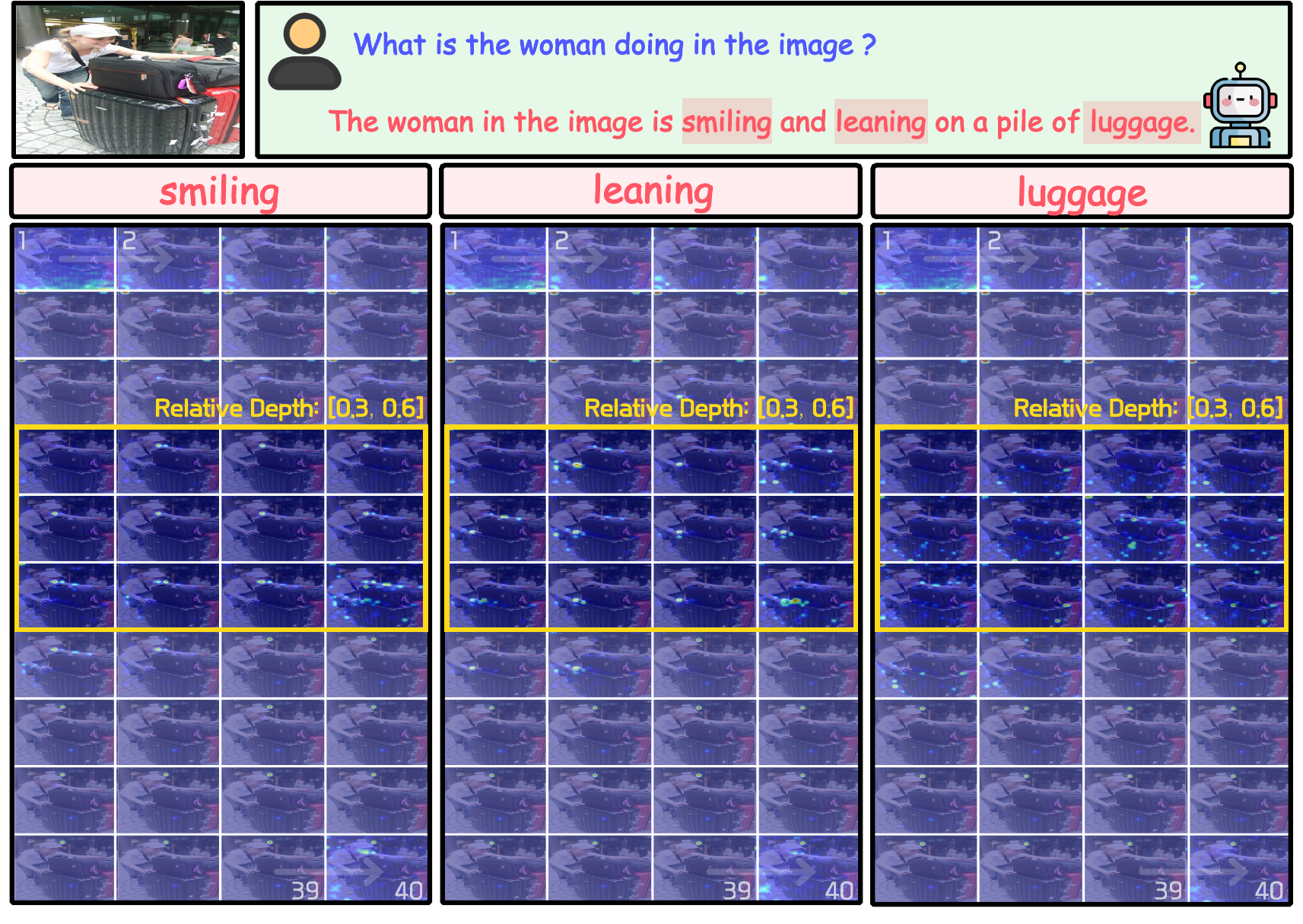}
    \caption{\textbf{Token-level attention adaptation for action versus object tokens.} During the generation step of the token \textit{`smiling'}, the model's attention is sharply localized around the woman's mouth. At the decoding step for \textit{`leaning'}, the focus correctly shifts to the woman's hands actively pushing the suitcases. In contrast, while predicting the object token \textit{`luggage'}, the attention distribution broadens to cover the large pile of suitcases spread across the scene.}
    \label{fig:obs2_luggage}
\end{figure}

This section provides additional examples (Figures \ref{fig:obs2_luggage}, \ref{fig:obs2_bird} and \ref{fig:obs2_baseball}) to support the attention granularity analysis discussed in the main paper. To systematically examine this phenomenon, we focus on relative depth, a $[0, 1]$ normalized depth coordinate representing the sequential position of decoder layers. Following the observation in the main paper that cross-modal interactions are most pronounced at intermediate depths, we analyze the relative depth range of $[0.3, 0.6]$. The yellow bounding boxes in the figures highlight this intermediate region, where Response-to-Vision attention is most prominent. 

Within this intermediate region, cross-modal attention patterns exhibit significant redundancy across adjacent decoder layers. However, when examined at the token level, these distributions reveal highly dynamic and granular grounding behaviors that strictly depend on the semantic role of the generated token.

The following examples demonstrate that response tokens require varying degrees of spatial focus. While some tokens necessitate sharp, highly localized attention on specific anatomical features or objects, others demand a diffuse, scene-level understanding. These visualizations reinforce the necessity of token-level adaptive supervision, as aggregating attention across the entire response sequence would inevitably blur these distinct spatial signals.

\clearpage

\begin{center}
    \includegraphics[width=0.8\linewidth]{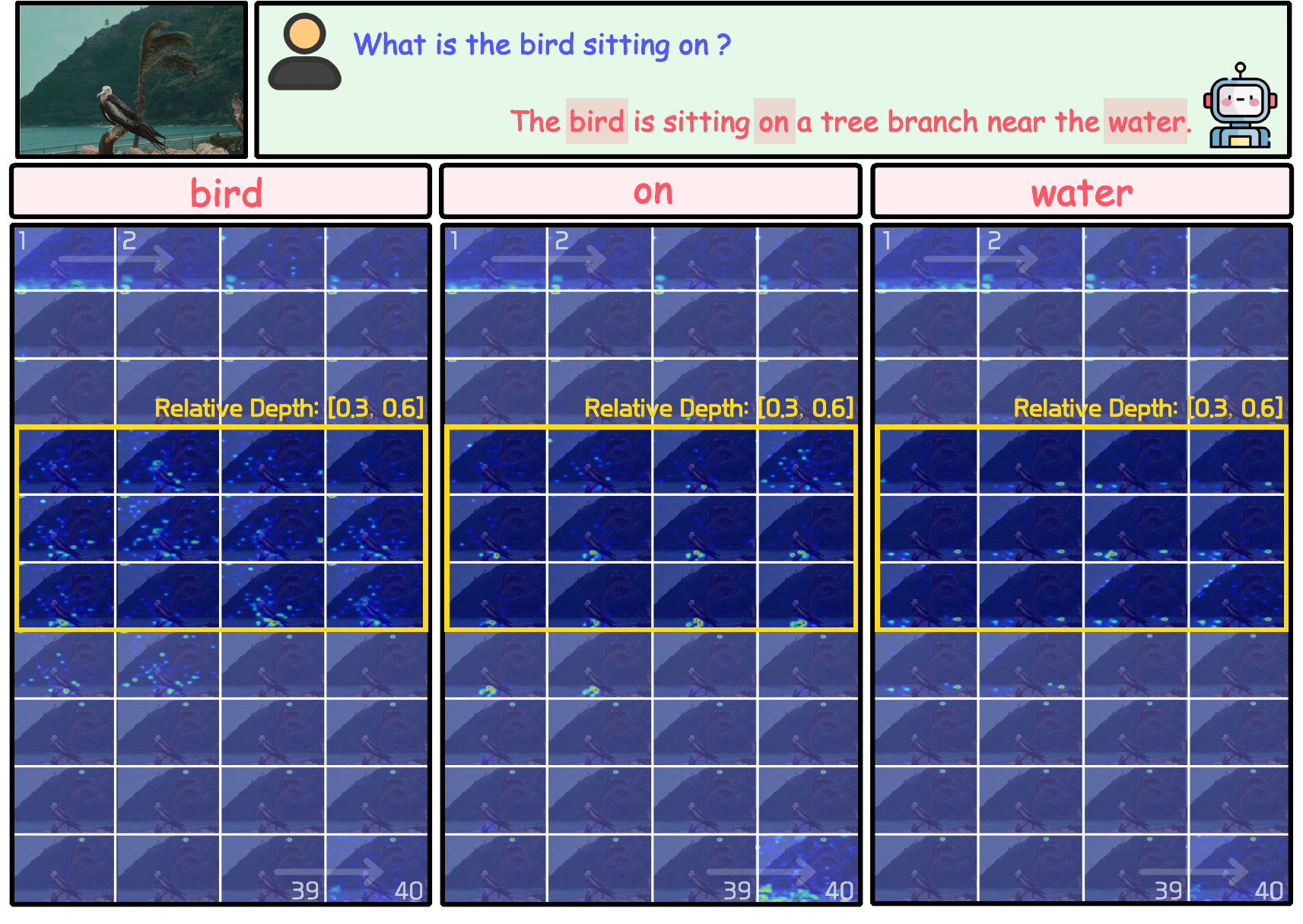}
    \captionof{figure}{\textbf{Attention dynamics for spatial relationships and entities.} During the generation of the token \textit{`bird'}, the attention map tightly localizes around the bird's body. At the decoding step for the spatial preposition \textit{`on'}, the model specifically attends to the bird's feet and the immediate branch structure it rests upon. When predicting \textit{`water'}, attention disperses evenly across the background water.}
    \label{fig:obs2_bird}

    \par\bigskip

    \includegraphics[width=0.8\linewidth]{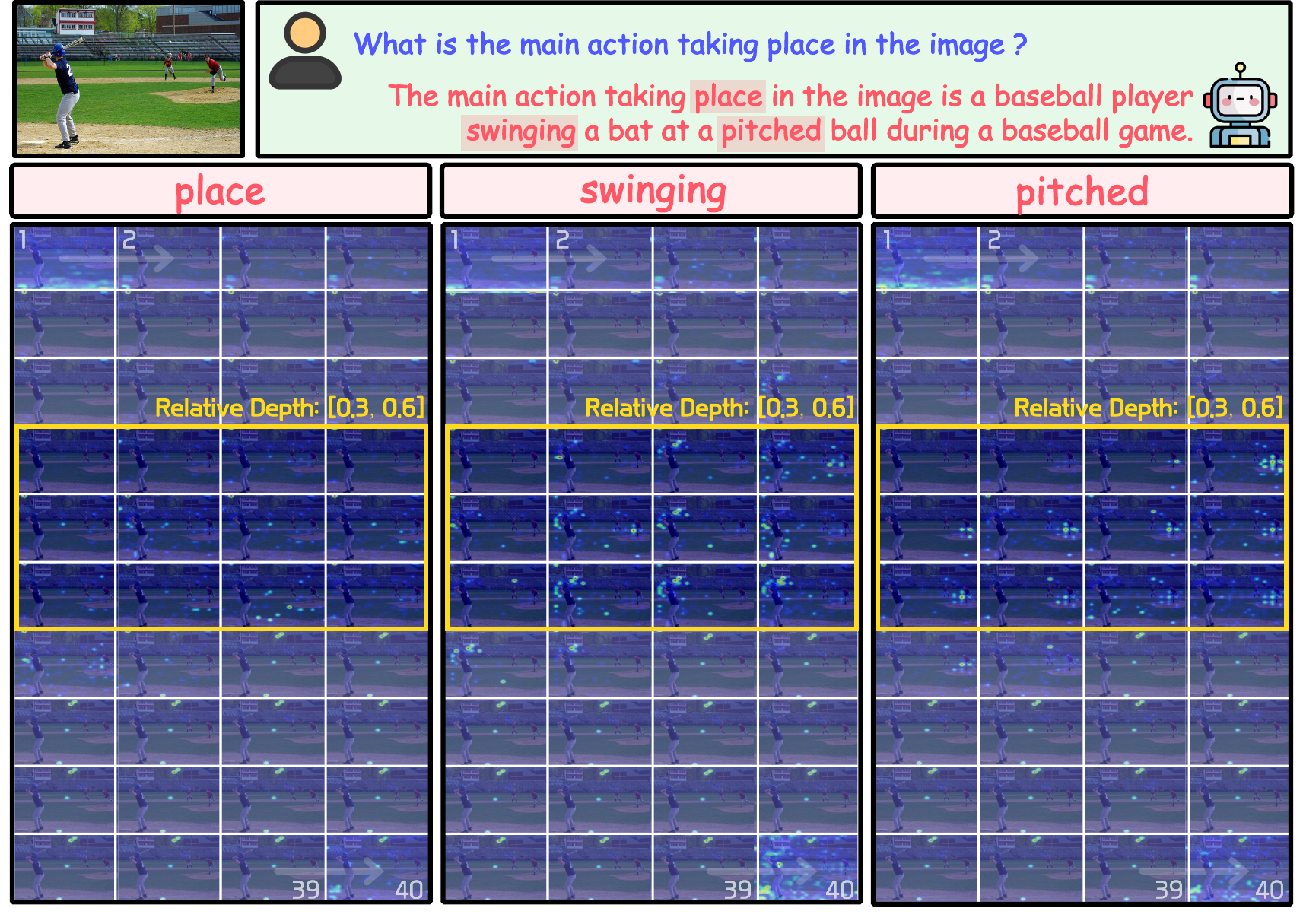}
    \captionof{figure}{\textbf{Variance in spatial focus between context-setting and action-specific tokens.} During the generation of the context-setting token \textit{`place'}, the model produces a diffuse, scene-wide attention pattern covering the baseball field. For action-specific tokens, the decoding steps exhibit high spatial precision: predicting \textit{`swinging'} heavily concentrates attention on the batter and the hands holding the bat, while generating \textit{`pitched'} sharply shifts the spatial focus to the pitcher standing on the mound.}
    \label{fig:obs2_baseball}
\end{center}

\clearpage

\section{Additional Experiments and Ablations}
\label{appendix:additional_results}

\subsection{Generalization to MobileLLaMA}
\label{appendix:mobilellama}

To examine whether \method~generalizes beyond the Qwen backbone family, we additionally evaluate it on MobileLLaMA-based MLLMs. As shown in Table~\ref{tab:mobilellama}, \method~substantially improves the MobileLLaMA-1.4B student over its PT-SFT baseline, with gains of $+3.0$ in VQA Avg and $+7.3$ in CR Avg. This confirms that the proposed response-time attention guidance is not restricted to a specific LLM backbone.

\begin{table}[h]
\centering
\caption{Experiments with the MobileLLaMA LLM backbone.}
\label{tab:mobilellama}
\renewcommand{\arraystretch}{0.9}
\setlength{\tabcolsep}{6pt}
\resizebox{0.8\linewidth}{!}{
\begin{tabular}{l|cc}
\toprule[0.15em]
\textbf{Setting} & \textbf{VQA $Avg$} & \textbf{CR $Avg$} \\
\midrule[0.1em]
MobileLLaMA-2.7B (PT-SFT) & 59.7 & 57.5 \\
MobileLLaMA-1.4B (PT-SFT) & 55.2 & 50.0 \\
\rowcolor{green}
\textbf{\method~2.7B $\rightarrow$ 1.4B} & \textbf{58.2} & \textbf{57.3} \\
\bottomrule[0.15em]
\end{tabular}
}
\end{table}

\subsection{Depth Range Ablation}
\label{appendix:depth_range_ablation}

Table~\ref{tab:depth_range_ablation} studies the effect of the relative depth range used for attention aggregation. The shifted intermediate range $[0.4, 0.7]$ achieves performance comparable to the default $[0.3, 0.6]$ range, whereas early and late ranges degrade both VQA and compositional reasoning performance. These results suggest that the benefit of \method~comes from supervising intermediate cross-modal grounding behavior rather than from a model-specific layer choice.

\begin{table}[h]
\centering
\caption{Ablation results on relative depth range for attention aggregation.}
\label{tab:depth_range_ablation}
\renewcommand{\arraystretch}{0.9}
\setlength{\tabcolsep}{6pt}
\resizebox{0.8\linewidth}{!}{
\begin{tabular}{l|cc}
\toprule[0.15em]
\textbf{Relative Depth Range} & \textbf{VQA $Avg$} & \textbf{CR $Avg$} \\
\midrule[0.1em]
Early $[0.0, 0.3]$ & 58.7 & 69.2 \\
Late $[0.6, 1.0]$ & 59.2 & 73.2 \\
Intermediate $[0.4, 0.7]$ (Shifted) & \textbf{60.3} & 75.0 \\
\rowcolor{green}
\textbf{Intermediate $[0.3, 0.6]$ (Default)} & 60.2 & \textbf{75.2} \\
\bottomrule[0.15em]
\end{tabular}
}
\end{table}

\subsection{Granularity of Response Supervision}
\label{appendix:response_granularity}

Table~\ref{tab:response_granularity} studies the granularity of response-token supervision. Aggregating attention targets over the entire response sequence yields lower performance than applying per-token response supervision. This indicates that token-wise supervision better preserves the step-specific grounding signals required during response generation.

\begin{table}[h]
\centering
\caption{Ablation results on response-token granularity.}
\label{tab:response_granularity}
\renewcommand{\arraystretch}{0.9}
\setlength{\tabcolsep}{6pt}
\resizebox{0.8\linewidth}{!}{
\begin{tabular}{c|cc}
\toprule[0.15em]
\textbf{Response Token Usage} & \textbf{VQA $Avg$} & \textbf{CR $Avg$} \\
\midrule[0.1em]
Aggregation & 57.7 & 70.4 \\
\rowcolor{green}
\textbf{Token-Level (Ours)} & \textbf{60.2} & \textbf{75.2} \\
\bottomrule[0.15em]
\end{tabular}
}
\end{table}

\section{Detailed Formulation of the Training Objective}
\label{appendix:loss_detail}
This section provides a comprehensive breakdown of the objective function optimized during the distillation stages of \method. While recent work has explored perception-decomposed confidence rewards for vision-language reasoning~\cite{pdcr}, \method~uses teacher-guided distillation to supervise response-time visual grounding. The overall training loss is formulated as the sum of three distinct components:

\begin{equation}
\mathcal{L}_{\text{\method}}
=
\mathcal{L}_{\text{LM}}
+
\mathcal{L}_{\text{uni\text{-}modal}}
+
\mathcal{L}^{(\text{\method})}_{\text{cross\text{-}modal}}.
\label{eq:appendix_total_loss}
\end{equation}

\subsubsection{Language Modeling Loss ($\mathcal{L}_{\text{LM}}$).}
We apply the standard autoregressive cross-entropy loss over the target response sequence to maintain the foundational generation capabilities of the student model:

\begin{equation}
\mathcal{L}_{\text{LM}}
=
-
\sum_{t=1}^{L_R}
\log p^{\mathcal{S}}
\!\left(
y_t \mid \mathbf{v}, \mathbf{x}, \mathbf{y}_{<t}
\right),
\label{eq:appendix_lm_loss}
\end{equation}
where $\mathbf{v}$, $\mathbf{x}$, and $\mathbf{y}$ denote the visual tokens, prompt tokens, and target response tokens, respectively.

\subsubsection{Uni-modal Knowledge Distillation Loss ($\mathcal{L}_{\text{uni-modal}}$).}
The uni-modal distillation term aligns the teacher and student models within their respective modalities. It combines a response-level logit matching term with an auxiliary vision-modality loss, denoted as $\mathcal{L}_{\text{aux}}$:

\begin{equation}
\mathcal{L}_{\text{uni-modal}} = \sum_{t=1}^{L_R} D_{KL} \left( p^{\mathcal{T}}(\cdot \mid \mathbf{v}, \mathbf{x}, y_{<t}) \parallel p^{\mathcal{S}}(\cdot \mid \mathbf{v}, \mathbf{x}, y_{<t}) \right) + \mathcal{L}_{\text{aux}}.
\label{eq:appendix_unimodal_main}
\end{equation}
The first term minimizes the Kullback-Leibler (KL) divergence between the teacher and student next-token probability distributions at each decoding step.

To ensure stable training and facilitate a fair comparison, the auxiliary loss $\mathcal{L}_{\text{aux}}$ strictly follows the configuration established in LLaVA-KD~\cite{llavakd}. It comprises vision-token logit matching and visual affinity matching:
\begin{equation}
\mathcal{L}_{\text{aux}}
=
\lambda_\alpha
\sum_{i=1}^{N}
D_{KL}\!\left(
p^\mathcal{T}_{\mathrm{vision},i}
\,\|\,
p^\mathcal{S}_{\mathrm{vision},i}
\right)
+
\lambda_{\beta}
\mathcal{D}_{\cos}
\!\left(
\mathbf{R}^{\mathcal{T}}_{\text{vision}},
\mathbf{R}^{\mathcal{S}}_{\text{vision}}
\right).
\label{eq:appendix_laux}
\end{equation}
Here, $p^\mathcal{T}_{\mathrm{vision},i}$ and $p^\mathcal{S}_{\mathrm{vision},i}$ denote the teacher and student output distributions for the $i$-th visual token, respectively, while $\mathbf{R}^{\mathcal{T}}_{\text{vision}}$ and $\mathbf{R}^{\mathcal{S}}_{\text{vision}}$ represent their corresponding visual affinity matrices. $\mathcal{D}_{\cos}$ denotes the cosine distance. We retain the original hyperparameter weighting from LLaVA-KD, setting $\lambda_{\alpha} = 1.0$ and $\lambda_{\beta} = 0.5$.

\subsubsection{Cross-modal Knowledge Distillation Loss ($\mathcal{L}_{\text{cross-modal}}$).}
The core contribution of our distillation framework is encapsulated in the cross-modal attention alignment term. As formulated in the main paper, the \method~objective dynamically balances the forward and reverse KL divergences based on the token-specific entropy of the teacher's attention distribution:
\begin{equation}
\mathcal{L}^{(\text{\method})}_{\text{cross-modal}}
=
\sum_{i \in \mathcal{I}_R}
\left[
\lambda_i
D_{KL}
\!\left(
\tilde{\mathbf{a}}^{\mathcal{T}}_i
\,\|\, 
\tilde{\mathbf{a}}^{\mathcal{S}}_i
\right)
+
(1-\lambda_i)
D_{KL}
\!\left(
\tilde{\mathbf{a}}^{\mathcal{S}}_i
\,\|\, 
\tilde{\mathbf{a}}^{\mathcal{T}}_i
\right)
\right].
\label{eq:appendix_crossmodal}
\end{equation}

\section{Limitations and Future Work}
\label{appendix:limitation_futurework}
While \method~effectively enhances cross-modal grounding, we identify two primary limitations in the current framework. 

First, the proposed distillation objective inherently assumes vocabulary consistency between the teacher and the student. Because the token-wise attention alignment and the response-level logit supervision are defined over aligned target response positions under teacher forcing, the current framework cannot be directly applied to teacher-student pairs with mismatched tokenizers or entirely disjoint vocabularies.

Second, although our method focuses on the response phase, the cross-modal supervision is computed under a standard teacher forcing paradigm using fixed ground-truth target sequences. Consequently, the distilled attention patterns are not explicitly optimized to dynamically adapt to the student’s own autoregressive generation trajectory, which may introduce a discrepancy between training-time supervision and inference-time behavior.

Despite this theoretical limitation, the main paper shows that \method~maintains high attention fidelity during autoregressive generation without teacher forcing, reaching approximately 0.72 cosine similarity at intermediate layers. While such high fidelity demonstrates that our approach meaningfully mitigates the train-inference mismatch, an inherent structural discrepancy inevitably persists.

Extending \method~to an on-policy distillation setting would therefore be a natural direction for future work~\cite{vgs}. By defining supervision along the student’s actual decoding trajectory and dynamically querying the teacher’s attention for student-generated tokens, such a formulation could provide a more faithful approximation of the cross-modal grounding behavior required at inference time.

\section{Reproducibility Statement}
\label{appendix:reproduce}
To ensure the full reproducibility of our proposed framework, \method, we provide comprehensive methodological details throughout the main paper. Furthermore, exhaustive training configurations and hyperparameters are specified in Appendix~\ref{appendix:implementation}, and the exact mathematical formulations of all objective functions are detailed in Appendix~\ref{appendix:loss_detail}.

\section{Qualitative Results}
\label{appendix:qual_sample}
\begin{figure}[p]
    \centering
    \includegraphics[width=1\linewidth]{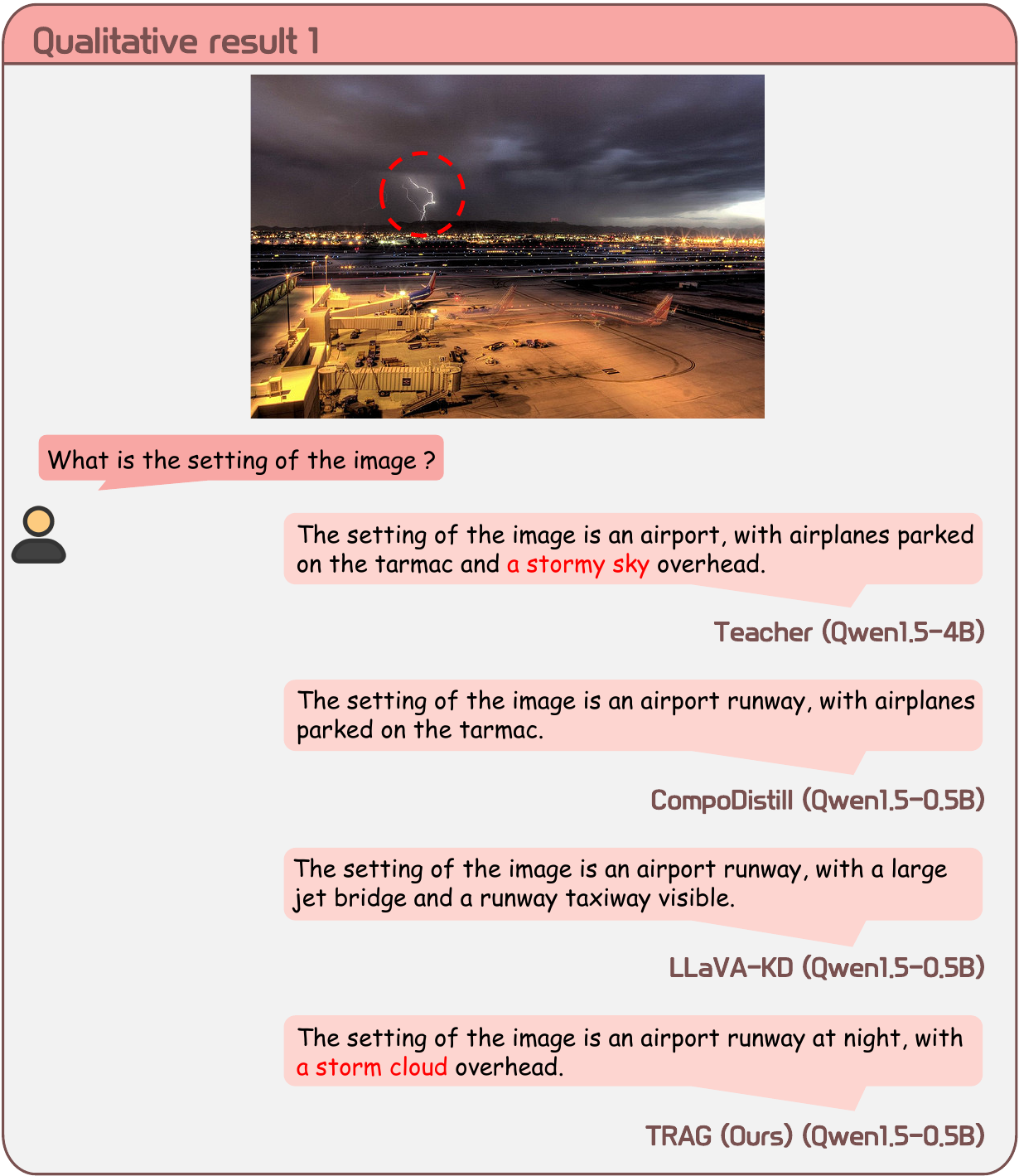}
    \caption{\textbf{Qualitative comparison on scene description and salient feature grounding.} For the question regarding the setting of the image, the teacher model provides a detailed response that captures both the overall scene context (``an airport'') and a salient localized attribute (``a stormy sky''). \method~produces a response that more closely reflects the teacher's response pattern during generation, correctly identifying the localized feature as ``a storm cloud.'' This example illustrates that response-phase attention distillation helps the student better align with the teacher's response characteristics while preserving fine-grained visual grounding.}
    \label{fig:inference1}
\end{figure}

\begin{figure}[p]
    \centering
    \includegraphics[width=1\linewidth]{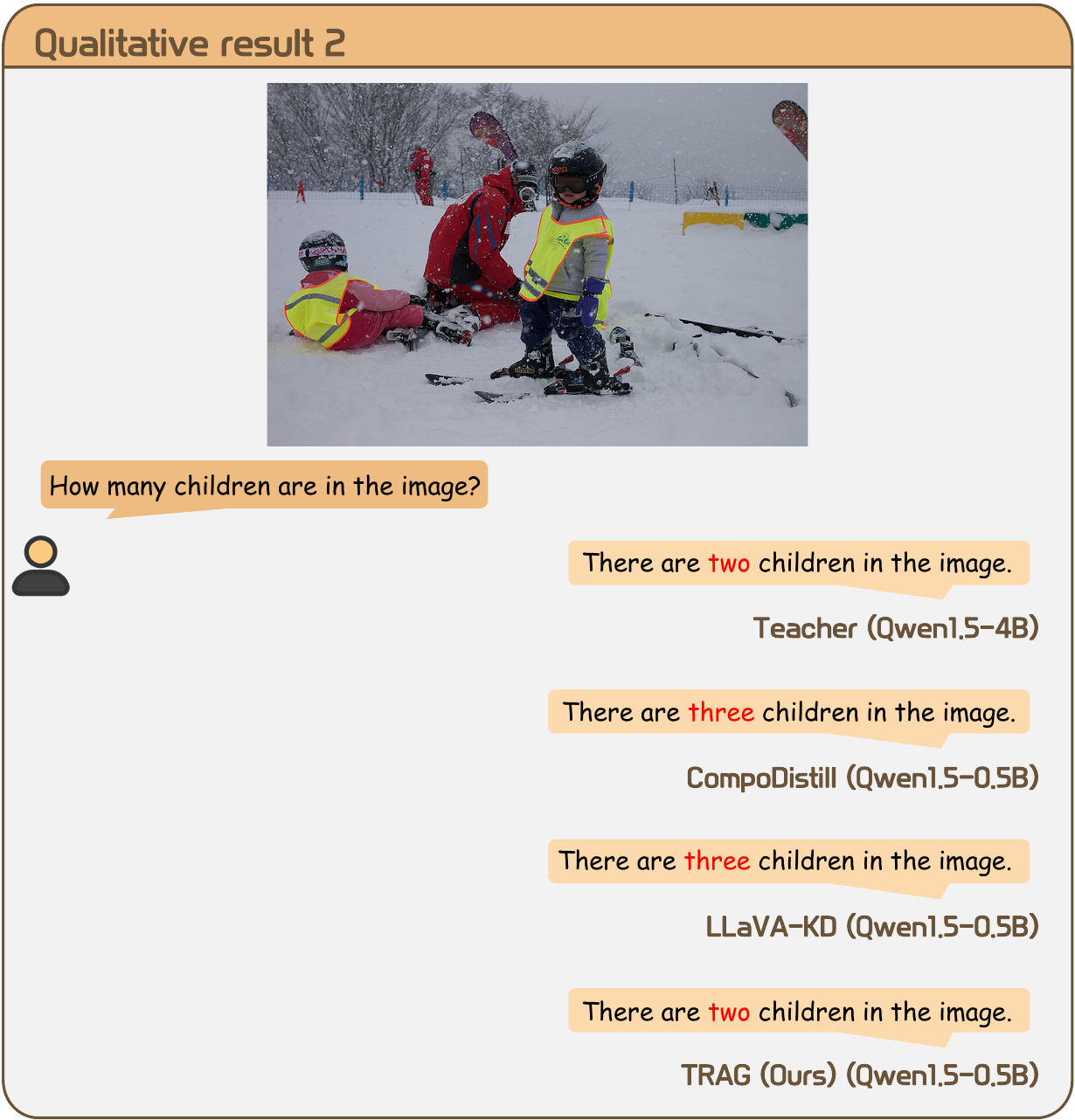}
    \caption{\textbf{Qualitative analysis of counting and precise object identification.} The foreground of the image prominently features two children and one adult, while an additional adult is visible in the background. In response to the question regarding the number of children, both the teacher model and our \method~correctly answer ``two'' children. Baselines LLaVA-KD and CompoDistill overcount by one, misidentifying the adult as a child. This discrepancy highlights that \method~possesses superior grounding capabilities, derived from response-attention distillation, to accurately distinguish object categories compared to existing state-of-the-art distillation frameworks.}
    \label{fig:inference2}
\end{figure}

\begin{figure}[p]
    \centering
    \includegraphics[width=1\linewidth]{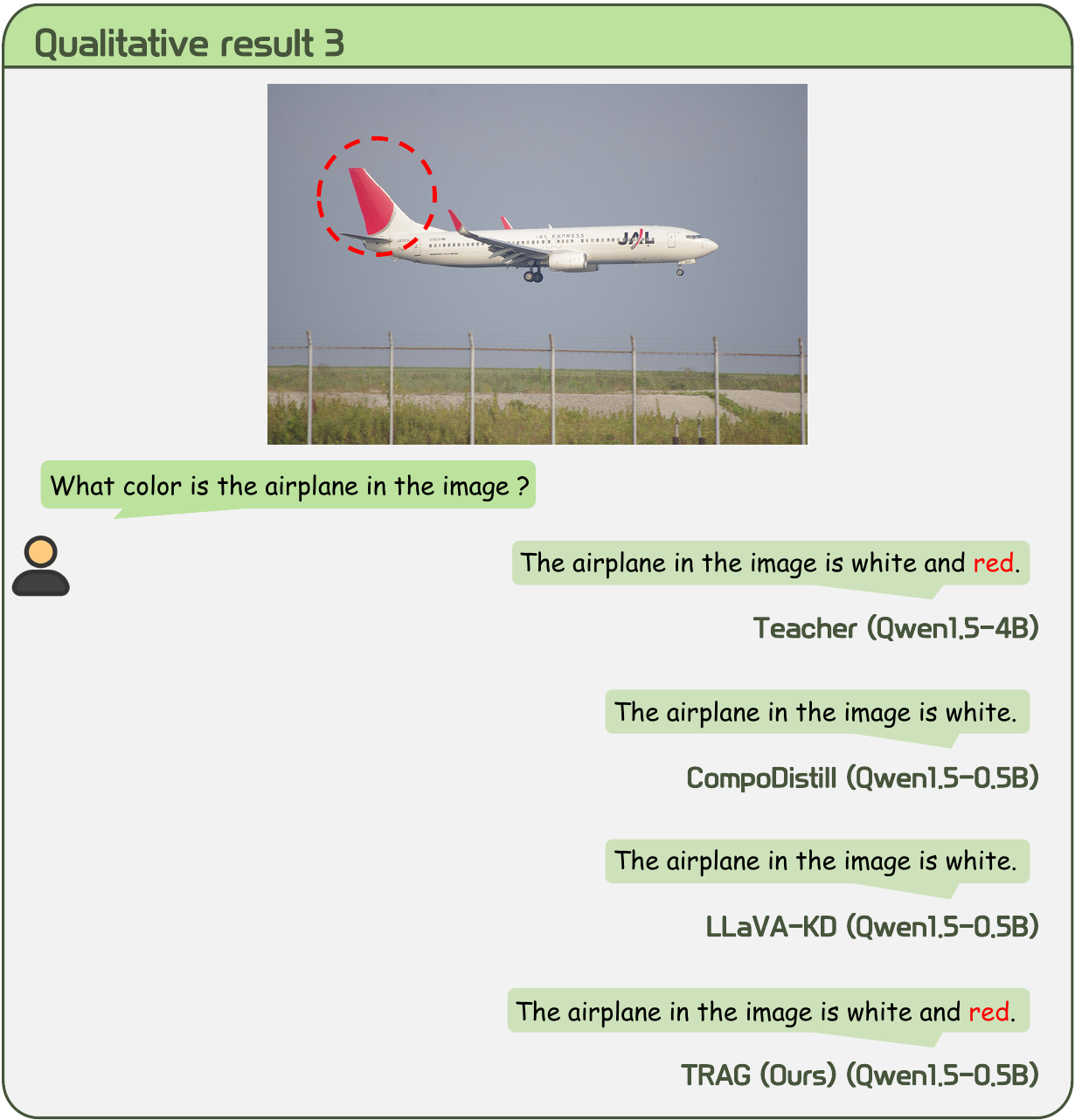}
    \caption{\textbf{Demonstration of capturing fine-grained object attributes.} When asked about the airplane's color, the teacher model captures both the predominant color (``white'') and the fine-grained, localized attribute (``red'' on the tail). Consistent with prior behavior, baseline methods only report the general predominant attribute (``white''). \method~demonstrates an aligned behavior with the teacher, successfully identifying the local ``red'' attribute, which reinforces that distilling attention during the response phase is effective for transferring the teacher's fine-grained visual grounding capabilities.}
    \label{fig:inference3}
\end{figure}

\begin{figure}[p]
    \centering
    \includegraphics[width=1\linewidth]{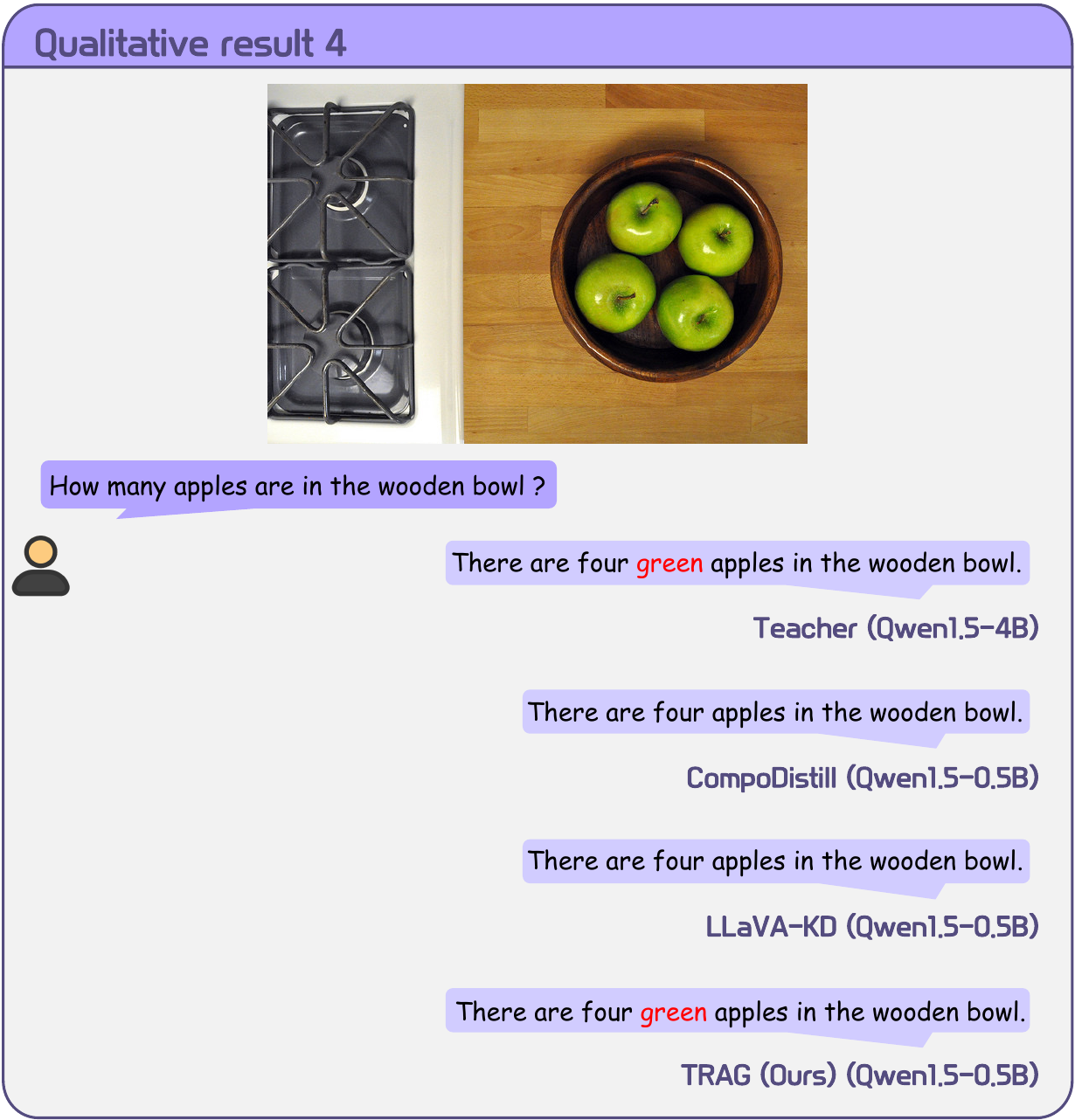}
    \caption{\textbf{Qualitative evidence for replicating descriptive response patterns.} For the question asking how many apples are in the bowl, all methods correctly answer ``four.'' Notably, both the teacher and \method~include the visually grounded attribute ``green'' in the response, even though it is not explicitly requested by the question. This example indicates that \method~more closely reflects the teacher's response characteristics beyond the minimum required answer.}
    \label{fig:inference4}
\end{figure}

In this section, we present a qualitative comparison of inference samples generated by our student model, \method~(Qwen1.5-0.5B), the teacher model (Qwen1.5-4B), and competitive baselines (CompoDistill and LLaVA-KD, both Qwen1.5-0.5B). Figures~\ref{fig:inference1}--\ref{fig:inference4} illustrate that \method~more closely matches the teacher's response behavior and fine-grained visual grounding at inference time. These examples support the effectiveness of our response-phase attention distillation in transferring the teacher’s response characteristics, allowing the student to move beyond generalized descriptions and capture salient localized features that are often overlooked by the baselines.

\clearpage

\clearpage

\clearpage

\clearpage

\end{document}